%% file: tcjepa_icml26_camera_ready.tex
\documentclass{article}

\usepackage{microtype}
\usepackage{graphicx}
\usepackage{subcaption}
\usepackage{booktabs} % for professional tables

\usepackage{hyperref}

\usepackage{url}
\input{math_commands.tex}

\usepackage{amsfonts}
\usepackage{nicefrac}       % compact symbols for 1/2, etc.
\usepackage{xcolor}         % colors
\usepackage{xspace}
\usepackage{multirow}
\usepackage{enumitem}
\usepackage{tabularx}
\usepackage{array}
\usepackage{colortbl}

\newcommand{\eg}{\emph{e.g.}}
\newcommand{\ie}{\emph{i.e.}}
\newcommand{\etc}{\emph{etc}}
\graphicspath{{./}}
\newcommand{\ourmethod}{TC-JEPA\xspace}
\newcommand{\ijepa}{I-JEPA\xspace}
\definecolor{lightgray}{rgb}{0.9, 0.9, 0.9}

\usepackage[accepted]{icml2026}

\usepackage{amsmath}
\usepackage{amssymb}
\usepackage{mathtools}
\usepackage{amsthm}

\usepackage[capitalize,noabbrev]{cleveref}

\theoremstyle{plain}

\theoremstyle{definition}

\theoremstyle{remark}

\usepackage[textsize=tiny]{todonotes}

\icmltitlerunning{\ourmethod}

\begin{document}

\twocolumn[
  \icmltitle{Text-Conditional JEPA for Learning Semantically Rich Visual Representations}



  \icmlsetsymbol{equal}{*}

  \begin{icmlauthorlist}
    \icmlauthor{Chen Huang}{comp}
    \icmlauthor{Xianhang Li}{comp}
    \icmlauthor{Vimal Thilak}{comp}
    \icmlauthor{Etai Littwin}{comp}
    \icmlauthor{Josh Susskind}{comp}
  \end{icmlauthorlist}

  \icmlaffiliation{comp}{Apple, Cupertino, United States}

  \icmlcorrespondingauthor{Chen Huang}{chen-huang@apple.com}

  \icmlkeywords{TC-JEPA, ICML}

  \vskip 0.3in
]



\printAffiliationsAndNotice{}  % no special notice (required even if empty)

\begin{abstract}

Image-based Joint-Embedding Predictive Architecture (\ijepa) offers a promising approach to visual self-supervised learning through masked feature prediction. However with the inherent visual uncertainty at masked positions, feature prediction remains challenging and may fail to learn semantic representations. In this work, we propose Text-Conditional JEPA (\ourmethod) that uses image captions to reduce the prediction uncertainty. Specifically, we modulate the predicted patch features using a fine-grained text conditioner that computes sparse cross-attention over input text tokens. With such conditioning, patch features become predictable as a function of text, thus are more semantically meaningful. We show \ourmethod improves downstream performance and training stability, with promising scaling properties. \ourmethod also offers a new vision-language pretraining paradigm based on feature prediction only, outperforming contrastive methods on diverse tasks, especially those requiring fine-grained visual understanding and reasoning.

\end{abstract}

\section{Introduction}

% where visual representations can be readily used on various downstream tasks.
Significant advances have been made in the field of Self-Supervised Learning (SSL) from images, with two common families of SSL approaches. Invariance-based methods produce high-level semantic representations by learning invariances across
augmented image views~\citep{chen2020big,BYOL,caron2021emerging}. However, some image augmentations may hurt downstream generalization on tasks that require different invariances~\citep{xiao2021what,mast}. Masked Image Modeling (MIM) methods require less prior knowledge and learn visual representations by reconstructing masked image patches in the pixel~\citep{MaskedAutoencoders} or latent space~\citep{baevski22a}. MIM typically learns patch-level features that prioritize local structure, and latent MIM methods like Image-based Joint-Embedding Predictive Architecture (\ijepa)~\citep{assran2023self} gain more popularity due to their ability to capture both local and semantic information.

\begin{figure}[!t]
\begin{center}
\vskip -0.1in
\centerline{\includegraphics[width=1.0\linewidth]{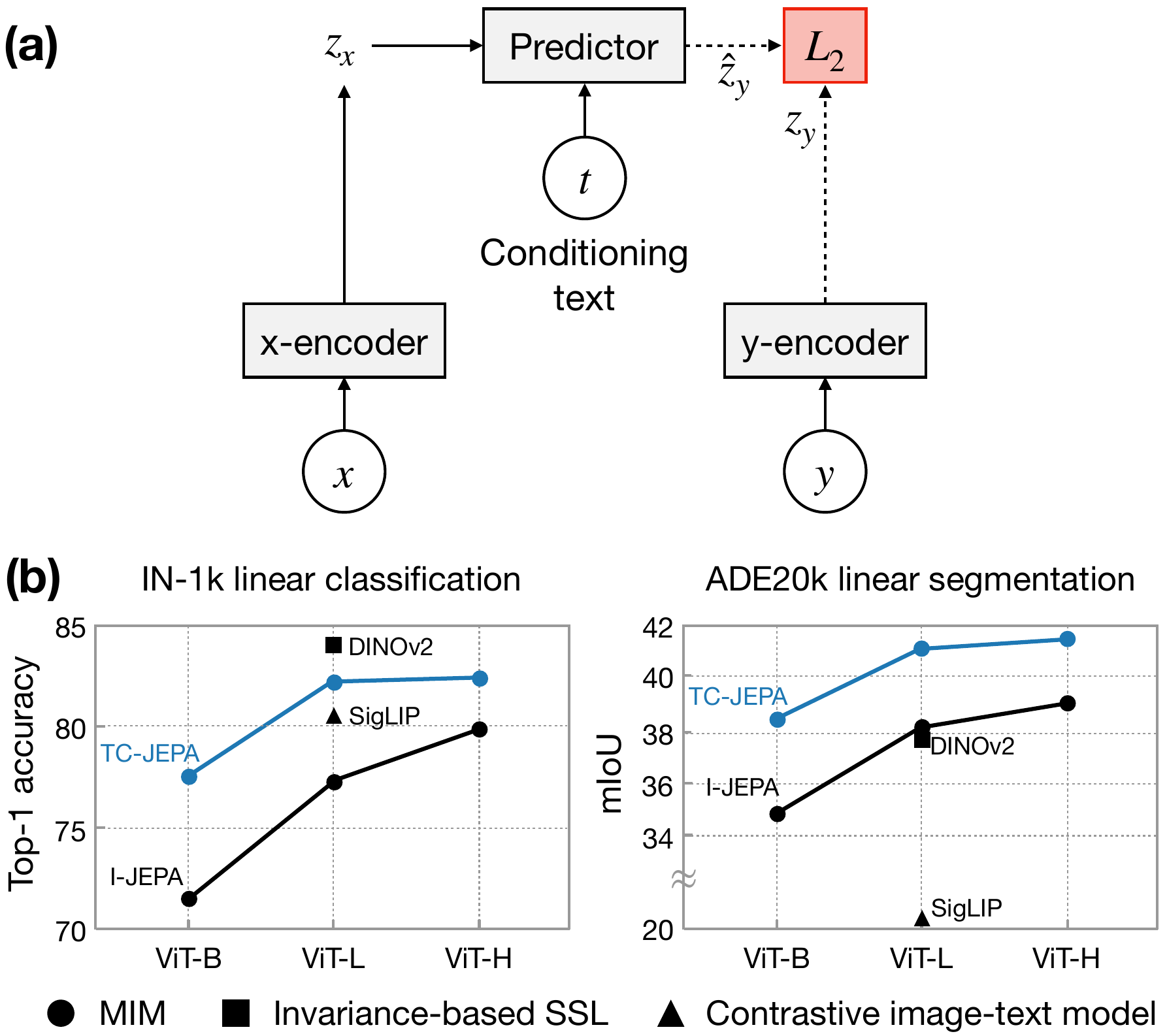}}
\end{center}
\vskip -0.25in
\caption{\textbf{(a)} \ourmethod is trained to predict the representation of a signal $y$ from that of signal $x$, using a predictor conditioned on text input $t$ to facilitate prediction. \textbf{(b)} \ourmethod vs. 3 types of visual representation learning methods: MIM (\ijepa), invariance-based SSL (DINOv2) and contrastive image-text training (SigLIP) methods. Note SigLIP is trained on a large dataset WebLI, while others are trained on the much smaller IN-21k dataset; both \ourmethod and SigLIP use weak text supervision from image captions. \ourmethod performs best for fine-grained image understanding (segmentation), scaling well with model size, and approaches the classification performance of SOTA invariance learning approach DINOv2 that requires handcrafted augmentations.} 
\label{fig:fig1}
\vskip -0.19in
\end{figure}

Recent works either improve the JEPA objective~\citep{darcet2025capi} or even scale to videos~\citep{bardes2024revisiting}. Despite these successes, JEPA's core pretext task still poses challenges. Namely, predicting features in arbitrary masked positions involves large uncertainties. For example, it is hard to predict the masked bookshelf in the dog image in Fig.~\ref{fig:method}, since a clean wall would also be a plausible prediction. The intrinsic prediction uncertainty often makes \ijepa sensitive to the masking strategy. When the mutual information between the context and masked patches is too low, feature prediction becomes challenging and may lead to representation collapse with no useful semantics encoded. Two recent attempts to address this issue use position-conditional encoders~\citep{littwin2024enhancing} or stochastic positional embeddings~\citep{barstochastic}, but the prediction difficulty remains without adding new information.

We propose to aid feature prediction using human- or LMM-generated image captions (required only for feature pretraining, not at test time). Intuitively, a caption about scene composition (\eg,~dog + bookshelf) can reveal the spatial relationships between context and target (bookshelf) windows. When the feature predictor is augmented with such text information, we can substantially reduce the feature prediction uncertainty. Hence we propose Text-Conditional JEPA (\textbf{\ourmethod}) that combines the predictive power of JEPAs with text conditioning inputs (Fig.~\ref{fig:fig1}), which is under-explored in the context of visual representation learning.
This way, we turn \ijepa into a text-conditional representation learner, where patch representations are now predictable or transformable when ``prompted'' by text, thus are more language aligned and semantically meaningful.

For strong text conditioning, we introduce a new fine-grained text conditioner. Specifically, we modulate the predicted patch features at multiple layers of the predictor, by using cross-attention over text tokens. This helps to identify fine-grained correspondences between image patches and word tokens, which is optimized in a self-supervised way to best facilitate feature prediction. We further discuss useful regularizations on the patch-word similarities using sparsity and consistency constraints. Overall, our \ourmethod is found to achieve improved performance on various tasks of classification and dense prediction, while being highly scalable and stable to train.

Note the use of image-text pairs in \ourmethod makes it comparable to language-supervised methods. One popular language-supervised method is CLIP~\citep{radford2021learning} that contrasts image and text features. However, CLIP tends to focus on high-level semantics and abstract away detailed information, hence struggling with fine-grained image understanding tasks.
Recent methods~\citep{li2021grounded,CLOC} improve by using grounding data,~\eg,~bounding box and region descriptions. Alternatives operate in an unsupervised way and incorporate fine-grained loss like the local-to-global image consistency loss~\citep{MaskCLIP,SILC}. More related to our method is unsupervised correspondence learning between image patches and text~\cite{yao2022filip,SPARC,DreamLIP}. One key difference is that \ourmethod learns such fine-grained correspondences for feature prediction rather than contrastive learning. In summary, we provide a new fine-grained vision-language pretraining method using the feature prediction objective, with no grounding data or contrastive loss.
Experiments show \ourmethod can achieve stronger performance than contrastive methods on different tasks, including dense prediction (\eg,~segmentation) and multimodal (image captioning and VQA) tasks.% like semantic  that require fine-grained spatial representations.

Our \textbf{main contributions} are as follows: 1) We propose \ourmethod to improve \ijepa via fine-grained text conditioning, which in turn produces semantically rich visual representations. 2) \ourmethod leads to improved downstream performance, training stability and scaling properties. 3) \ourmethod offers a fine-grained vision-language pretraining paradigm based on feature prediction only, which outperforms contrastive methods on diverse (fine-grained) tasks.

% encode fine-grained details without any grounding data, which is more scalable. The idea is to enrich the learning signal from image-text pairs
% which is stable to train, and exhibits promising scaling properties.

\section{Related Work}

\textbf{Visual SSL} has evolved along two different paths. Invariance-based methods encourage the similarity between augmented views of the same image, using either a contrastive~\citep{chen2020big} or non-contrastive~\citep{BYOL} loss with different mechanisms to prevent collapse. MoCo v3~\citep{chen2021mocov3} and DINO~\cite{caron2021emerging} use similar ideas for training with ViT~\citep{dosovitskiy2020vit}. These methods often excel at high-level vision tasks like classification, but are limited by the carefully designed data augmentations~\citep{xiao2021what,mast}.
Whereas, MIM methods learn visual representations through a more generic pretext task: reconstructing masked image parts in pixel~\citep{MaskedAutoencoders} or latent space~\citep{baevski22a,assran2023self}. While MIM typically captures local image information, latent MIM methods strike a better balance between learning local and highly semantic representations. There is also iBOT~\citep{zhou2021ibot} and DINOv2~\citep{oquab2024dinov} that combines invariance learning and MIM, and Franca~\citep{venkataramanan2025franca} and Web-DINO~\citep{fan2025scaling} are the recent scaling efforts of iBOT and DINOv2, respectively.

\textbf{JEPA} offers a promising approach to latent MIM that learns by predicting masked information in feature space. JEPA has been successfully applied to audio~\citep{baevski22a}, image~\citep{assran2023self} and video data~\citep{bardes2024revisiting}. A recent line of work attempts to improve the JEPA prediction task. For example, CAPI~\citep{darcet2025capi} predicts latent clusterings as the target representations to stabilize training. To address the prediction uncertainty at masked positions, some remedies include using position-conditional encoders~\citep{littwin2024enhancing} or stochastic positional embeddings~\citep{barstochastic}. We propose to use text information to explicitly reduce uncertainties, which was not explored for JEPA prediction before.

\textbf{Language-supervised methods} are popularized by contrastive models like CLIP~\citep{radford2021learning} and SigLIP~\citep{10377550} that learn joint embeddings for image and text on large-scale datasets~\citep{xu2024demystifying}.
Subsequent efforts improve the image-text alignment by using captioning loss~\citep{yu2022coca}, masked language modeling~\citep{li2022blip} or a cross-modal encoder via self-attention~\citep{ALBEF,li2022blip}. Alternative works aim at training fine-grained vision-language models, either by using extra grounding data~\citep{li2021grounded,CLOC} or incorporating fine-grained loss like local-to-global image consistency~\citep{MaskCLIP,SILC} and unsupervised correspondence learning between image patches and text tokens~\cite{yao2022filip,SPARC} or entire text captions~\citep{DreamLIP}. We similarly learn such unsupervised correspondences but for JEPA prediction instead, leading to a non-contrastive, fine-grained vision-language pretraining approach.

\section{Method}

% We describe the I-JEPA baseline~\citep{assran2023self} in Section~\ref{sec:ijepa}, and then our Text-Conditional JEPA (\ourmethod) method in Section~\ref{sec:tcjepa}.
% with a focus on an image-grounded text conditioner. We also discuss feature-level pooling to fuse information from multiple text conditioning inputs.

\subsection{\ijepa Baseline}
\label{sec:ijepa}

The \ijepa objective is to predict the representations of masked image parts,~\ie,~target patches $y=\{y_j|j\in B_y\}$, given the context patches $x=\{x_i|i\in B_x\}$ in the same image, where $B_{x}$ and $B_{y}$ denote the set of context and target indices respectively. A multi-block masking strategy is used to produce non-overlapping patches $x$ and $y$.

\textbf{Context and target encoding.} The context patch representations $z_x=\{z_{x_i}\}_{i\in B_x}$ are extracted via an encoder $f_{\theta}$ parameterized using standard ViT~\citep{dosovitskiy2020vit}: $z_{x_i}=f_{\theta}(x_i)$ where each $z_{x_i}\in \mathbb{R}^{d}$ has a position embedding $p_{i} \in \mathbb{R}^{d}$ added to it. The groundtruth target patch features $z_{y_j}=f_{\bar\theta}(y_j) \in \mathbb{R}^{d}$ are processed by $f_{\bar\theta}$ that is an exponential moving average of $f_{\theta}$.

\textbf{Target prediction.} We predict target features $\hat z_{y}=\{\hat z_{y_j}\}_{j\in B_y}$ via a predictor $g_{\phi}$ implemented as a narrow ViT. Specifically, $\hat z_{y} = g_{\phi}(z_x,m)$,~\ie,~the predictor takes as input all the context features $z_{x}$ and, conditioned on a sequence of positional mask tokens $m=\{m_j\}_{j\in B_y}$, predicts the target representations at patch positions specified by the mask tokens. Here, every mask token $m_j \in \mathbb{R}^{d}$ is the sum of the position embedding $p_{j}$ of the $j^{th}$ target patch and a shared learnable vector $\tilde m$,~\ie,~$m_j=p_{j}+\tilde m$.
% \begin{equation}
%     m_j=p_{j}+\tilde m.
% \label{eq1}
% \end{equation}

\textbf{Loss.} The encoder $f_{\theta}$ and predictor $g_{\phi}$ are trained simultaneously by minimizing the feature prediction error:
\begin{equation}
    \mathcal{L}_{\text{predict}}= \frac{1}{|B_y|} \sum_{j\in B_y} \| \hat z_{y_j} - z_{y_j}\|_2,
\label{eq1}
\end{equation}
where $z_{y_j}=f_{\bar\theta}(y_j)$ uses both an exponential-moving average feature extractor and a stop-gradient operation to prevent representation collapse.

\subsection{Text-Conditional JEPA}
\label{sec:tcjepa}

We propose to aid JEPA prediction by conditioning the predictor $g_{\phi}$ on (synthesized) \textit{image captions} in addition to the positional mask tokens. Image captions are particularly helpful in case of low mutual information between the context and target patches, which can benefit from text descriptions often about scene composition or object interactions.

For conditioning purposes, given a text caption $c$ associated with input image, we first map $c$ using the pretrained T5~\citep{2020t5} into a sequence of word embeddings $t=[t_1,\dots,t_S] \in \mathbb{R}^{d_t \times S}$, where $t_s \in \mathbb{R}^{d_t}$ and $S$ is the sequence length. We use the language model T5 rather than CLIP's text encoder~\citep{radford2021learning} since T5 can better represent the composition and order information in natural language captions~\citep{yuksekgonul2023when}.

\begin{figure*}[!t]
\begin{center}
\vskip -0.1in
\centerline{\includegraphics[width=1.0\linewidth]{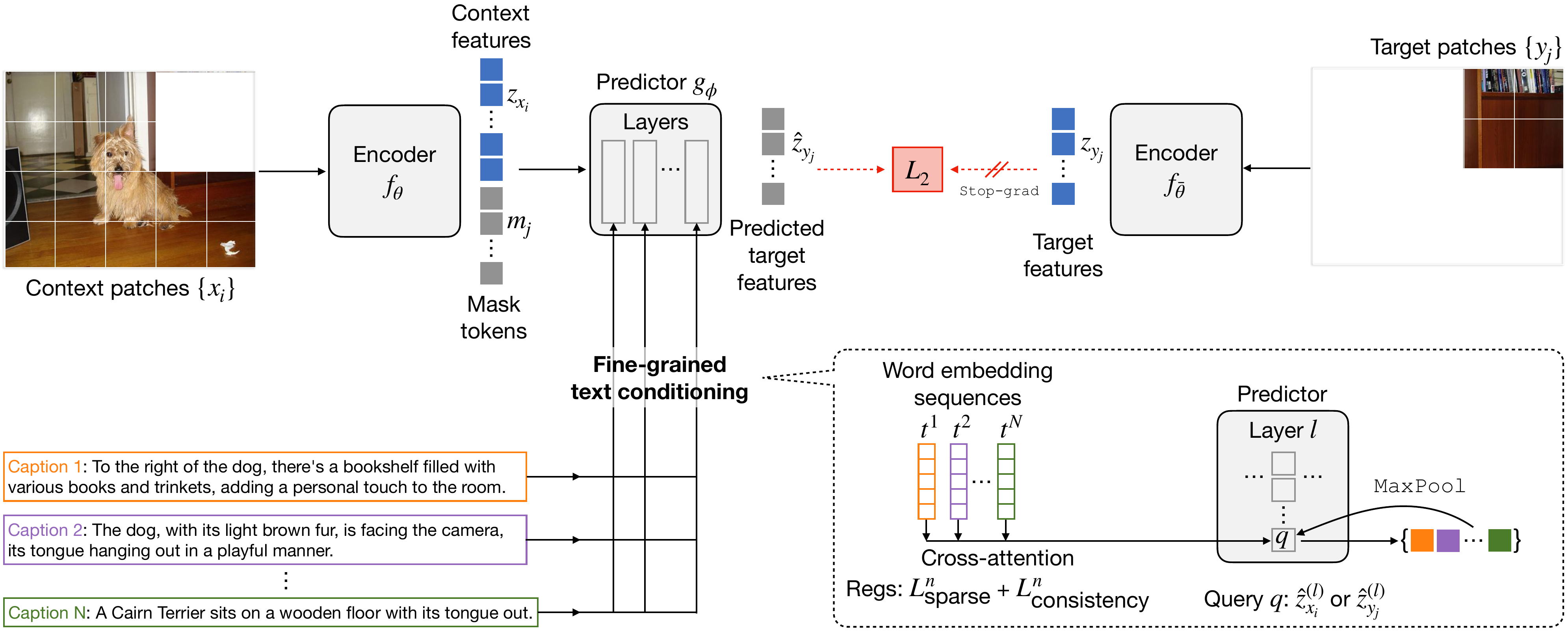}}
\end{center}
\vskip -0.18in
\caption{\textbf{\ourmethod}: conditioning the I-JEPA predictor $g_{\phi}$ on text captions using a fine-grained text conditioner. Conditioning is applied to the patch features predicted at multiple layers of $g_{\phi}$, using cross attention over the word embedding sequences $\{t^1,\dots,t^N\}$ extracted for $N$ captions. This leads to multi-caption-conditioned patch features that are then max-pooled at each layer. Our text conditioning process is akin to \textit{self-supervised} visual grounding, which identifies the fine-grained patch-word correspondences that are only optimized for target feature prediction. To further improve the self-supervised process, for conditioning with each ($n^{th}$) caption, we impose sparsity constraint $\mathcal{L}_{\text{sparse}}^{n}$ and cross-layer consistency constraint $\mathcal{L}_{\text{consistency}}^{n}$ on the patch-word similarities.}
\label{fig:method}
\vskip -0.02in
\end{figure*}
% $g_{\phi}$ predicts target patch features $\{\hat z_{y_j}\}$ given the context patch features $\{ z_{x_i}\}$ obtained via encoder $f_{\theta}$ and positional mask tokens $\{m_j\}$.
% Such grounded text conditioning makes patch features more reliably predicable between local image regions as a function of text, leading to highly semantic patch representations.

To condition predictor $g_{\phi}$ on the extra sequence $t$, one straightforward way is simply to append $t$ as additional tokens to the input sequence of $g_{\phi}$,~\ie,~$g_{\phi}(z_x,m,t)$. This is often referred to as \textbf{sequence conditioning}~\citep{Garrido2024LearningAL,assran2025vjepa2}. However, sequence conditioning will increase the sequence length processed by the predictor ViT, which requires additional model capacity with non-negligible overhead in memory and compute. Moreover, such conditioning is only applied at the predictor's input layer, with its effect rapidly vanishing in deeper layers.

% Self-supervised sparse visual grounding
\textbf{Cross attention over word sequence.} To address the above drawbacks, we propose to condition predictor $g_{\phi}$ by using lightweight cross attention over the word sequence at multiple layers of $g_{\phi}$. We choose to directly condition the multi-layer patch representations in $g_{\phi}$, because this allows us to compute patch-word similarities (\ie,~fine-grained image-text correspondences) that unlock capabilities akin to visual grounding\footnote{Visual grounding~\citep{11235566} is a task of identifying the fine-grained correspondence between words in a text caption and objects/patches in an image.}. 
Concretely, we define query $q\in \mathbb{R}^d$ as each of the patch features predicted at the $l^{th}$ layer of $g_{\phi}$ for $l\in [1,L]$,~\ie,~$q \in \{\hat z_x^{(l)},\hat z_y^{(l)}\}$. Then $q$ cross-attends to the word embedding sequence $t \in \mathbb{R}^{d_t \times S}$ giving:
\begin{eqnarray}
&&\!\!\!\!\!\!\!\!\!\!\!\!\!\!\!\!\text{Attention}(q^{(l)},K^{(l)},V^{(l)}) \!=\!\sum_{s=1}^S\text{softmax}\left( {q^{(l)}}^{\top}\cdot K^{(l)}_{:,s} \right)\cdot V^{(l)}_{:,s}, \notag \\
&&\!\!\!\!\!\!\!\!\!\!\!\!\!\!\!\!q^{(l)} =W^{(l)}_Q \cdot q, \;\; K^{(l)}=W^{(l)}_K \cdot t, \;\; V^{(l)}=W_V^{(l)} \cdot t, \notag \\
&&\!\!\!\!\!\!\!\!\!\!\!\!\!\!\!\!q\leftarrow q+\text{Attention}(q^{(l)},K^{(l)},V^{(l)}),
\label{eq2}
\end{eqnarray}
where $W^{(l)}_Q \in \mathbb{R}^{d \times d}$, $W^{(l)}_K \in \mathbb{R}^{d \times d_t}$ and $W^{(l)}_V \in \mathbb{R}^{d \times d_t}$ are the learnable query, key and value matrices at layer $l$. Each cross-attention layer is residual and its output is added back to $q$, followed by an MLP network and LayerNorm.

In doing so, predictor $g_{\phi}(z_x,m,t)$ is conditioned in a patch-specific way: text $t$ separately updates the features predicted for each context patch $\hat z_{x_{i|i\in B_x}}^{(l)}$ or target patch $\hat z_{y_{j|j\in B_y}}^{(l)}$. Also note our text conditioning is attentive, which enables a selective use of word tokens in $t$ for every patch, bearing similarity to visual grounding. However, one key difference from supervised visual grounding is that we find such fine-grained correspondences in a \textit{self-supervised} way, without any grounding annotations in given image-text pairs.
The only supervision we use is the patch feature prediction error (Eq.~(\ref{eq1})),~\ie,~we identify the word-patch correspondences that best support accurate feature prediction.

To further improve the unsupervised process as defined by Eq.~(\ref{eq2}), we regularize the cosine patch-word similarities $O^{(l)}_i=\max(\cos(q^{(l)},K^{(l)}),0) \in \mathbb{R}^{S}$ when they are positive,~\ie,~between semantically related patch-word pairs. Note $O^{(l)}_i$ is a rectified similarity vector computed between the $i^{th}$ patch for $i\in \{B_x,B_y\}$ and the entire word sequence $t$ at layer $l$. Then we impose a sparsity constraint on $O^{(l)}_i$ to maximize the selectivity of patch features with respect to related words. It is also found helpful to enforce cross-layer consistency of $O^{(l)}_i$ to obtain similar word selections across layers for each patch. We do so by penalizing the deviation between $O^{(l)}_i$ and the cross-layer mean $\bar{O}_i$: 
\begin{eqnarray}
&&\!\!\!\!\!\!\!\!\!\!\!\!\!\!\!\!\! \mathcal{L}_{\text{sparse}} \!=\! \frac{1}{|B_x|\!\!+\!\!|B_y|}  \!\sum_{i\in \{B_x,B_y\}} \!\!\frac{1}{L}\sum_{l=1}^L \|O^{(l)}_i\|_1, \; \bar{O}_i \!=\! \frac{1}{L}\sum_{l=1}^L O^{(l)}_i, \notag \\
&&\!\!\!\!\!\!\!\!\!\!\!\!\!\!\!\!\! \mathcal{L}_{\text{consistency}} \!=\! \frac{1}{|B_x|\!\!+\!\!|B_y|}  \!\sum_{i\in \{B_x,B_y\}} \frac{1}{L}\sum_{l=1}^L \|O^{(l)}_i - \bar{O}_i\|_1, \notag \\
&&\!\!\!\!\!\!\!\!\!\!\!\!\!\!\!\!\!\mathcal{L}= \mathcal{L}_{\text{predict}} + \lambda \mathcal{L}_{\text{sparse}} + \beta \mathcal{L}_{\text{consistency}},
\label{eq3}
\end{eqnarray}
where $\lambda$ and $\beta$ are the loss coefficients.

With the overall loss function, we obtain a fine-grained text conditioner that is jointly optimized for feature prediction. Such a conditioner can reliably provide language-aware modulation on image patch feature prediction. This is desirable from the perspective of visual representation learning, because it means feature representations are predictable or transformable between local image regions as a function of text. This lends our patch representations with some level of multimodal understanding capabilities, making them text-sensitive and semantically meaningful.

\textbf{Multi-caption conditioning.} When multiple text captions $\{c_n\}_{n \in [1,N]}$ are available to describe an image, they are more capable of capturing the richness of visual content; and with more text information, we may further improve the conditioning effect. Consider the image in Fig.~\ref{fig:method} where the context patches are insufficient to predict representations of the target patch, a random bookshelf in the background. Compared to using only one text caption, using multiple captions is more likely to cover the background element or its spatial relationship with foreground objects, which could better guide feature prediction for the target bookshelf.

% ,~\eg,~attributes of object parts vs. interactions with other objects.
For effective multi-caption conditioning, we need a good strategy to fuse text information.
Since different captions often capture different visual aspects, each caption offers a unique relationship between context and target patches. In other words, diverse captions often provide differing signals to condition the same inter-patch feature transformation. This makes it suboptimal to directly concatenate captions as a single text input to predictor $g_{\phi}$, since a target patch can simultaneously attend to multiple captions each with a different conditioning signal.

Here we choose to independently condition $g_{\phi}$ using each caption $c_n$, followed by a feature-level fusion strategy. Let $t^n$ denote the word sequence for $c_n$. The query patch features $q$ at the $l^{th}$ layer of $g_{\phi}$ will cross-attend to each $t^n$ using Eq.~(\ref{eq2}). We denote the separately conditioned patch features as $\hat z_{y_{j,n}}^{(l)}$ or $\hat z_{x_{i,n}}^{(l)}$ for $n \in [1,N]$, and the rectified cosine patch-word similarities as $O^{(l)}_{i,n}$. Then for the multi-caption-conditioned feature set $\{\hat z_{y_{j,n}}^{(l)} \}$ or $\{\hat z_{x_{i,n}}^{(l)} \}$ at layer $l$, we max-pool them along the $n$ dimension to fuse the most useful text information from $N$ captions. The pooling result $\hat z_{y_j}^{(l)}$ or $\hat z_{x_i}^{(l)}$ is fed as input to the next layer.

We also regularize the patch-word similarities $\{O^{(l)}_{i,n}\}$ when conditioning the $i^{th}$ patch with all the $n\in[1,N]$ captions at layer $l$. Let $\mathcal{L}_{\text{sparse}}^{n}$ denote the sparsity loss computed using Eq.~(\ref{eq3}) for conditioning with the $n^{th}$ caption. Similarly, the consistency loss for the $n^{th}$ caption is denoted as $\mathcal{L}_{\text{consistency}}^{n}$. Then our overall training loss becomes:
\begin{equation}
    \mathcal{L}= \mathcal{L}_{\text{predict}} + \frac{\lambda}{N}  \sum_{n=1}^N\mathcal{L}_{\text{sparse}}^{n} + \frac{\beta}{N} \sum_{n=1}^N\mathcal{L}_{\text{consistency}}^{n}.
\label{eq4}
\end{equation}
% \begin{eqnarray}
% &&\!\!\!\!\!\!\!\!\!\!\!\!\!\!\! \mathcal{L}_{\text{sparse}} \!=\! \sum_{i\in \{B_x,B_y\}} \!\frac{1}{N}\sum_{n=1}^N \frac{1}{L}\sum_{l=1}^L \|O^{(l)}_{i,n}\|_1, \;\, \bar{O}_{i,n}\!=\!\frac{1}{L}\sum_{l=1}^L O^{(l)}_{i,n}, \notag \\
% &&\!\!\!\!\!\!\!\!\!\!\!\!\!\!\! \mathcal{L}_{\text{consistency}}= \sum_{i\in \{B_x,B_y\}} \frac{1}{N}\sum_{n=1}^N \frac{1}{L}\sum_{l=1}^L \|O^{(l)}_{i,n} - \bar{O}_{i,n}\|_1.
% \label{eq4}
% \end{eqnarray}

\section{Experimental Setup}

% demonstrate the benefits of introducing text priors into the JEPA-based SSL framework.
\subsection{Pretraining on ImageNet}
\textbf{Data preparation}. IN-1k and IN-21k~\citep{ImageNet} are the golden pretraining datasets for most visual SSL methods. We compare with recent SSL methods all pretrained on the same ImageNet dataset, either IN-1k or IN-21k, except that our \ourmethod uses the dataset enriched with synthetic image captions. Following~\citep{DreamLIP}, we use ShareGPT4V~\citep{chen2024sharegpt4v} to synthesize captions, with an average of 8.3/8.7 caption sentences per image for IN-1k/-21k. Appendix~\ref{sec:gen_cap} includes the caption examples, statistics and generation details.

\textbf{Implementation.} We use the ViT architecture of ViT-B/16, ViT-L/16 or ViT-H/14 as image encoder $f_{\theta}$, which is jointly trained with the predictor $g_{\phi}$ and text conditioner. For text conditioning purposes, we randomly sample $N$ (default: 8) captions from the synthesized ones for each image. Note $N$ is capped by the number of available synthetic captions per image. Loss coefficients in Eq.~(\ref{eq4}) are $\lambda=0.1$, $\beta=0.5$.
We include in Appendix~\ref{sec:additional_ablation} the analysis of hyperparameters as well as our \textbf{training stability}, while Appendix~\ref{sec:train_efficiency} discusses \textbf{compute cost}.
The full training and evaluation details are included in Appendix~\ref{sec:train_eval_details}.

\subsection{Pretraining on Image-Text Datasets}
\textbf{Data preparation}. To further scale up training, we leverage the large image-text datasets CC12M~\citep{changpinyo2021cc12m} and YFCC15M~\citep{YFCC100M}. This also allows fair comparisons with recent contrastive vision-language models pretrained using these two datasets. We follow~\citep{DreamLIP} again to use ShareGPT4V to enrich the image captions available on CC12M and YFCC15M (details in Appendix~\ref{sec:gen_cap}), which facilitates strong text conditioning in our \ourmethod. We combine the raw caption with ShareGPT4V-synthesized ones for each image, which are then randomly sampled ($N=8$) for our pretraining.

\textbf{Implementation.} We train the ViT-B/16 and ViT-L/14 models for $f_{\theta}$. Training and evaluation details are described in Appendix~\ref{sec:train_eval_details}. We use the same loss coefficients $\lambda=0.1$, $\beta=0.5$ as downstream performance is not very sensitive to them (Appendix~\ref{sec:additional_ablation}).
% Training on image-text datasets is also found to be scalable and efficient.

\begin{table}[!t]
% \vskip 0.2in
\caption{\textbf{Linear probing results on IN-1k}. All methods are pretrained on IN-1k images, and \ourmethod uses extra text supervision.}
\label{tb:lin_IN1k}
\begin{center}
\vskip -0.1in
\resizebox{0.9\linewidth}{!}{
\begin{tabular}{llll}
\toprule
Method & Arch. & Epochs & Top-1 \\
\midrule
\rowcolor{lightgray} \multicolumn{4}{l}{MIM methods (no augmentations)} \\
data2vec~\citep{baevski22a} & ViT-L/16 & 1600 & 77.3 \\ \midrule
\multirow{3}{*}{MAE~\citep{MaskedAutoencoders}}  & ViT-B/16 & 1600 & 68.0 \\
  & ViT-L/16 & 1600 & 76.0 \\
  & ViT-H/14 & 1600 & 77.2 \\ \midrule
\multirow{3}{*}{\ijepa~\citep{assran2023self}} & ViT-B/16 & 600 & 72.9 \\
  & ViT-L/16 & 600 & 77.5 \\
  & ViT-H/14 & 300 & 79.3 \\ \midrule
\multirow{3}{*}{StoP~\citep{barstochastic}} & ViT-B/16 & 600 & 74.5 \\
  & ViT-L/16 & 600 & 78.5 \\
  & ViT-H/14 & 300 & 79.6 \\ \midrule
\multirow{3}{*}{\ourmethod (ours)} & ViT-B/16 & 600 & 75.8 \\
  & ViT-L/16 & 600 & 79.6 \\
  & ViT-H/14 & 300 & \textbf{80.4} \\ 
\rowcolor{lightgray} \multicolumn{4}{l}{Invariance-based SSL methods (with augmentations)} \\
SimCLR v2~\citep{chen2020big} & RN152 (2$\times$) & 800 & 79.1 \\ \midrule
BYOL~\citep{BYOL} & RN200 (2$\times$) & 800 & 79.6 \\ \midrule
\multirow{2}{*}{MoCo v3~\citep{chen2021mocov3}}  & ViT-B/16 & 300 &  76.7 \\
  & ViT-BN-L/7 & 300 &  \textbf{81.0} \\ \midrule
\multirow{2}{*}{DINO~\citep{caron2021emerging}}  & ViT-B/16 & 400 &  78.1 \\
  & ViT-B/8 & 300 &  80.1 \\ \midrule
\multirow{2}{*}{iBOT~\citep{zhou2021ibot}}  & ViT-B/16 & 250 &  79.8 \\
  & ViT-L/16 & 250 &  \textbf{81.0} \\
\bottomrule
\end{tabular}
}
\end{center}
\vskip -0.3in
\end{table}

\section{Results}

\subsection{Evaluating ImageNet Pretraining}

\begin{table*}[!t]
\caption{\textbf{Transfer performance on other classification and dense tasks} (object detection and semantic segmentation). All methods are pretrained on IN-1k, and our \ourmethod uses extra text supervision. Linear: linear evaluation. FT: fine-tuning.}
\vskip -0.1in
\label{tb:IN1k_downstream}
\begin{center}
\resizebox{0.9\textwidth}{!}{
\begin{tabular}{lccccccccc}
\toprule
 &  & \multicolumn{3}{c}{Classification} & Detection & \multicolumn{4}{c}{Semantic segmentation (mIoU)} \\
\cmidrule(lr){3-5}
\cmidrule(lr){6-6}
\cmidrule(lr){7-10}
 & & CIFAR100 & Places205 & iNat18 & COCO & \multicolumn{2}{c}{ADE20k} & \multicolumn{2}{c}{VOC}\\
\cmidrule(lr){3-5}
\cmidrule(lr){6-6}
\cmidrule(lr){7-8}
\cmidrule(lr){9-10}
Method & ViT & \multicolumn{3}{c}{Linear probing accuracy} & AP$^{b}$ & Linear & FT & Linear & FT\\ \midrule
\rowcolor{lightgray} \multicolumn{10}{l}{MIM methods (no augmentations)} \\
data2vec~\citep{baevski22a} & L/16 & 81.6 & 54.6 & 28.1 & 52.9 & 38.8 & 54.4 & 59.1 & 75.3 \\
MAE~\citep{MaskedAutoencoders} & H/14 & 77.3 & 55.0 & 32.9 & 50.1 & 33.3 & 50.9 & 67.6 & 80.7 \\
\ijepa~\citep{assran2023self} & H/14 & 87.5 & 58.4 & 47.6 & 53.7 & 36.9 & 51.2 & 64.6 & 81.4 \\
StoP~\citep{barstochastic} & H/14 & 87.7 & 58.4 & 50.9 & 53.5 & 36.6 & 53.3 & 66.9 & 82.3 \\
\ourmethod (ours) & H/14 & \textbf{88.5} & \textbf{59.1} & \textbf{54.8} & \textbf{55.2} & \textbf{39.5} & \textbf{55.7} & \textbf{70.4} & \textbf{83.8} \\ \midrule
\rowcolor{lightgray} \multicolumn{10}{l}{Invariance-based SSL methods (with augmentations)} \\
DINO~\citep{caron2021emerging} & B/16 & 84.8 & 55.2 & \textbf{50.1} & 50.1 & 27.5 & 44.7 & 63.1 & 75.4 \\
iBOT~\citep{zhou2021ibot} & B/16 & \textbf{85.5} & 56.7 & 50.0 & 51.2 & 32.3 & 48.2 & 64.3 & 78.2 \\ \midrule
\ourmethod (ours) & B/16 & 84.5 & \textbf{56.9} & 49.3 & \textbf{52.6} & \textbf{36.6} & \textbf{51.6} & \textbf{66.2} & \textbf{81.5} \\
\bottomrule
\end{tabular}
}
\end{center}
\vskip -0.1in
\end{table*}

\textbf{Image classification.} Table~\ref{tb:lin_IN1k} shows linear probing results on IN-1k using the encoder pretrained on the same dataset. Several observations: 1) \ourmethod outperforms \ijepa across 3 model scales, demonstrating our capabilities of encoding high-level semantics in visual representations. The larger \ourmethod models also narrow the performance gap with invariance-based methods such as iBOT (80.4 vs. 81) without requiring hand-crafted data augmentations.
2) With text conditioning, \ourmethod consistently outperforms StoP that improves JEPA prediction via stochastic positional embeddings.
3) When compared to other MIM methods like MAE and data2vec, \ourmethod achieves significant gains in both performance and training speed (5$\times$ fewer epochs).

\textbf{Transfer learning.} To assess generalization, we perform linear probing on other classification datasets: CIFAR100~\citep{Krizhevsky09learningmultiple}, Places205~\citep{NIPS2014_3fe94a00} and iNat18~\citep{Horn_2018_CVPR}. Table~\ref{tb:IN1k_downstream} shows that \ourmethod outperforms previous MIM methods across the three datasets, and decreases the gap again with augmentation invariance-based methods, even surpassing them on Places205.

\begin{table}[!t]
% \vskip 0.2in
\caption{\textbf{Scaling pretraining data} to IN-21k and image-text datasets CC27M (YFCC15M+CC12M) to compare with the state of the arts. Our \ourmethod belongs to MIM methods but using weak text supervision. $\dagger$: distilled from DINOv2-G on LVD-142M (results cited from~\citep{venkataramanan2025franca}). C100: CIFAR100. Class.: linear classification. Seg.: linear segmentation (mIoU).}
\label{tb:compare_IN21k}
\begin{center}
\vskip -0.1in
\resizebox{1.0\linewidth}{!}{
\begin{tabular}{lp{0.2in}p{0.65in}p{0.2in}p{0.2in}c}
\toprule
 & & & \multicolumn{2}{c}{Class.} & Seg. \\
\cmidrule(lr){4-5}
\cmidrule(lr){6-6}
Method & ViT & Data & IN1k & C100 & ADE \\
\midrule
\rowcolor{lightgray} \multicolumn{6}{l}{MIM methods (no augmentations)} \\
\ijepa~\citep{assran2023self} & L/16 & IN-21k & 77.2 & 88.7 & 38.2 \\
CAPI~\citep{darcet2025capi} & L/14 & IN-21k & 79.3 & 89.2 & 38.7 \\
\ourmethod (ours) & L/16 & IN-21k & \textbf{82.1} & \textbf{91.6} & 41.2 \\
\ourmethod (ours) & L/16 & CC27M & 81.4 & 91.2 & \textbf{42.1} \\
\rowcolor{lightgray} \multicolumn{6}{l}{Invariance+MIM-based SSL methods (with augmentations)} \\
iBOT~\citep{zhou2021ibot} & L/16 & IN-21k & 82.3 & 92.8 & 39.2 \\
Franca~\citep{venkataramanan2025franca} & L/14 & IN-21k & 84.5 & \textbf{94.1} & 41.4 \\
DINOv2~\citep{oquab2024dinov} & L/14 & IN-21k & 84.0 & 93.9 & 37.8 \\
DINOv2~\citep{oquab2024dinov}$\dagger$ & L/14 & LVD-142M & \textbf{86.3} & 93.4 & \textbf{41.8} \\
Web-DINO~\citep{fan2025scaling} & L/16 & MC-2B & 82.4 & 90.7 & 40.3 \\
\rowcolor{lightgray} \multicolumn{6}{l}{Language-supervised methods} \\
SigLIP~\citep{10377550} & L/16 & WebLI & 80.5 & 89.8 & 20.5 \\
SigLIP2~\citep{tschannen2025siglip} & L/16 & WebLI & \textbf{82.5} & \textbf{91.2} & \textbf{24.6} \\
\bottomrule
\end{tabular}
}
\end{center}
\vskip -0.2in
\end{table}

\begin{figure}[!t]
\begin{center}
% \vskip -0.1in
\centerline{\includegraphics[width=1.0\linewidth]{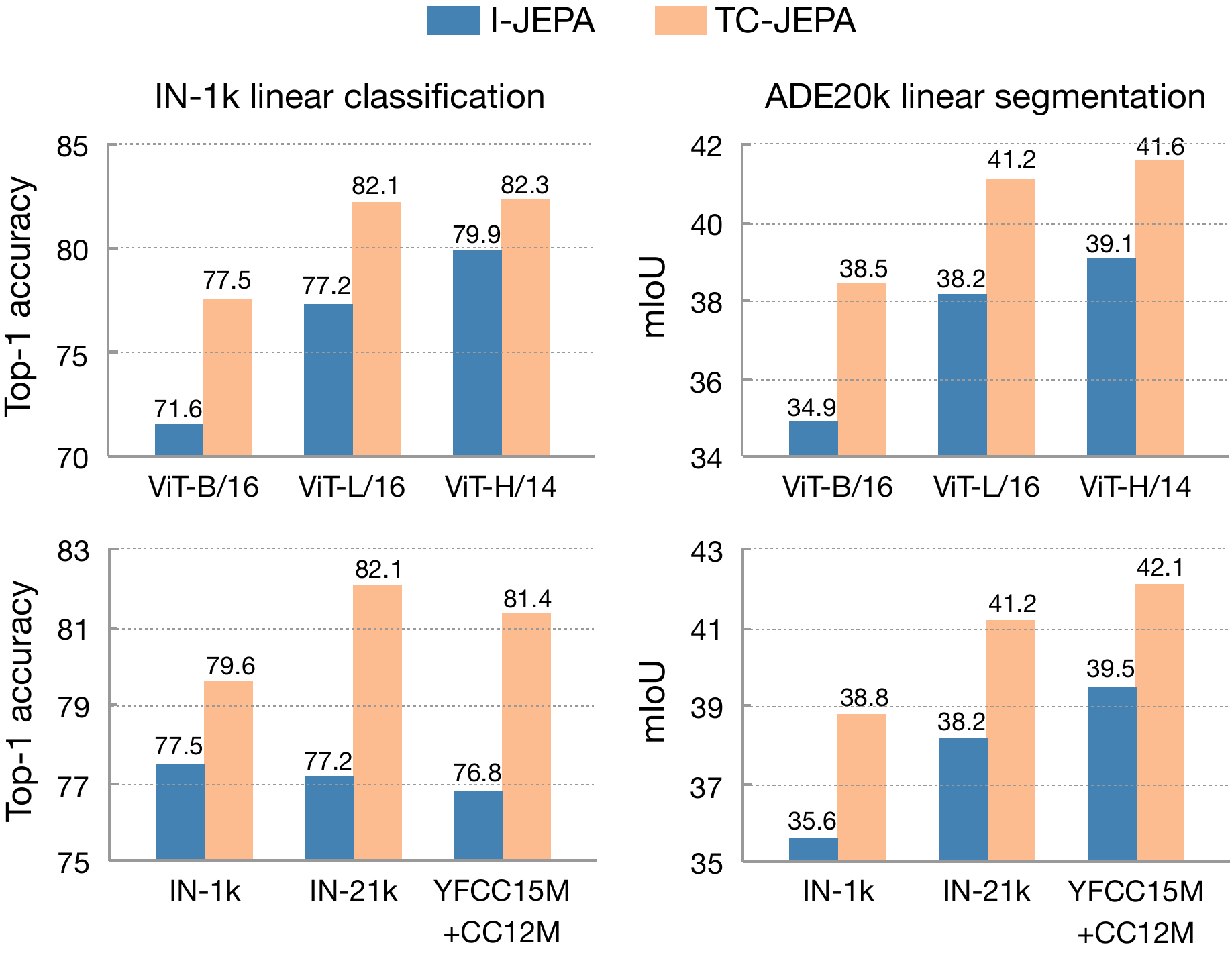}}
\end{center}
\vskip -0.3in
\caption{\textbf{Scaling behavior of \ijepa and \ourmethod \textit{w.r.t.} both model and training data size}. Top row: scaling up model size when trained on IN-21k. Bottom row: increasing pretraining data when training ViT-L/16.} 
\label{fig:jepa_scaling}
\vskip -0.2in
\end{figure}

\textbf{Dense prediction tasks.} We consider object detection on COCO dataset~\citep{LinMBHPRDZ14} and semantic segmentation on ADE20k~\citep{zhou2017scene} and Pascal VOC~\citep{Pascalvoc}, and report AP$^{b}$ and mIoU for the two tasks. These tasks require good localization and recognition, which could benefit from spatial and semantic patch representations. Note segmentation is performed under both linear and full fine-tuning protocols, with the former being a more explicit evaluation of representation quality. 

In Table~\ref{tb:IN1k_downstream}, we see the local patch features learned by our \ourmethod significantly improve over invariance-based methods on dense tasks, since the latter prioritize global image features. For instance, on COCO detection task, \ourmethod achieves 2.5\% and 1.4\% AP$^{b}$ gains over DINO and iBOT respectively, suggesting limited localization capabilities with invariance learning. When compared with prior MIM methods that similarly learn local features, \ourmethod outperforms them across different tasks and evaluation settings by learning text-sensitive, and hence more semantically meaningful patch features. Note \ourmethod achieves this without performance degradation on classification tasks, highlighting its ability to produce strong visual representations that can capture both global and fine-grained information.

\textbf{Scaling up training data.} We further explore the scaling behavior of \ourmethod \textit{w.r.t.} data size. After scaling from IN-1k to IN-21k (13M images) and to the combined image-text datasets YFCC15M+CC12M, we compare with state-of-the-art self- and text-supervised models in Table~\ref{tb:compare_IN21k}.

We find our IN-21k pretrained model already matches or exceeds the classification performance of the text-supervised models SigLIP/v2 trained on a much larger dataset WebLI. In the meantime, \ourmethod achieves huge gains on ADE20k segmentation (\eg,~+16.6\% mIoU over SigLIP2). This confirms the advantage of our text-conditional JEPA objective over contrastive image-text alignment, which tends to dismiss image details that are key to dense tasks.

In comparison to MIM methods, \ourmethod significantly outperforms the \ijepa and recent CAPI methods on both classification and segmentation tasks. Methods of iBOT, Franca and DINOv2 further combine MIM and invariance learning for stronger SSL. When trained on the same IN-21k dataset, \ourmethod and Franca achieve the highest mIoU on ADE20k (41.2/41.4), and \ourmethod remains competitive on classification tasks (within 2.5\% of top performers). After scaling training data to YFCC15M+CC12M, \ourmethod sets a new state of the art (42.1) on ADE20k. Notably on ADE20k, \ourmethod surpasses DINOv2 (41.8) distilled from LVD-142M and Web-DINO (40.3) trained on MC-2B, while using 5$\times$ and 75$\times$ less data respectively.

Finally, Fig.~\ref{fig:jepa_scaling} shows \ourmethod scales well with both data and model size, outperforming \ijepa baseline in all settings. Interestingly, increasing pretraining data for \ijepa does not exhibit a clear scaling trend on IN-1k classification.

\subsection{Evaluating Pretraining with Image-Text Datasets}

\begin{table}[!t]
% \vskip 0.2in
\caption{\textbf{Comparing with contrastive vision-language models pretrained on image-text datasets}. CC27M denotes our combined YFCC15M+CC12M datasets. The SPARC and DreamLIP methods use the same extra synthetic text captions as \ourmethod. $\dagger$: our implementation. Class.: linear classification. Det.: object detection (AP$^{b}$). Seg.: semantic segmentation (mIoU) via supervised finetuning transfer.}
\label{tb:compare_IT_methods}
\begin{center}
\vskip -0.1in
\resizebox{1.0\linewidth}{!}{
\begin{tabular}{lp{0.2in}p{0.6in}p{0.2in}p{0.25in}c}
\toprule
 & & & Class. & Det. & Seg. \\
% \cmidrule(lr){4-5}
% \cmidrule(lr){6-6}
Method & ViT & Data & IN1k & COCO & ADE \\
\midrule
CLIP~\citep{radford2021learning} & B/16 & YFCC15M & 66.5 & 43.6 & 47.8 \\
BLIP~\citep{li2022blip} & B/16 & Merged14M & 71.2 & 43.1 & 46.9 \\
MaskCLIP~\citep{MaskCLIP} & B/16 & YFCC15M & 73.7 & 45.4 & 50.5 \\
SPARC~\citep{SPARC}$\dagger$ & B/16 & YFCC15M & 73.4 & 52.0 & 52.3 \\
DreamLIP~\citep{DreamLIP} & B/16 & YFCC15M & 75.2 & 47.2 & 49.6 \\
\ourmethod (ours) & B/16 & YFCC15M & \textbf{77.1} & \textbf{54.5} & \textbf{55.2} \\ \midrule
GroupViT~\citep{xu2022groupvit} & S/16 & CC27M & 69.8 & 44.3 & 50.1 \\
SPARC~\citep{SPARC}$\dagger$ & B/16 & CC27M & 74.1 & 52.6 & 54.0 \\
DreamLIP~\citep{DreamLIP} & B/16 & CC30M & 78.6 & 50.7 & 52.4 \\
\ourmethod (ours) & B/16 & CC27M & 77.3 & 55.6 & 56.8 \\ %\midrule
\ourmethod (ours) & L/14 & CC27M & \textbf{81.9} & \textbf{58.0} & \textbf{58.8} \\
\bottomrule
\end{tabular}
}
\end{center}
\vskip -0.2in
\end{table}

Pretraining on image-text datasets enables comparing with popular vision-language models mostly trained via contrastive image-text alignment (\eg,~CLIP). Note our \ourmethod offers a non-contrastive vision-language pretraining method, based on text-conditional, latent MIM. \ourmethod further captures fine-grained image-text correspondence (between patches and text tokens) under this new paradigm. To evaluate whether \ourmethod improves fine-grained image understanding, we again test on classification and other tasks that require such fine-grained capabilities. We focus on comparing with recent contrastive methods using the same/similar image-text datasets and training backbones. Special attention is paid to contrastive methods that use a fine-grained loss but no grounding data.

\textbf{Classification and dense tasks.} Table~\ref{tb:compare_IT_methods} shows that \ourmethod, when trained on YFCC15M, significantly outperforms CLIP across tasks by up to 10.9\%, highlighting our ability to learn both semantically meaningful and spatially precise features. On the other hand, BLIP combines contrastive image-text learning with masked language modeling, which improves classification performance over CLIP but hurts localization (detection and segmentation). MaskCLIP further improves by adding a local-to-global image consistency loss, but underperforms our approach. SPARC and DreamLIP are more related to \ourmethod in that they similarly model fine-grained correspondence between image patches and multiple words or text captions. We differ by learning such properties in a non-contrastive way and perform better across the board.

When scaling to larger datasets YFCC15M+CC12M, our \ourmethod still surpasses SPARC and DreamLIP on dense tasks while achieving classification performance comparable to DreamLIP (77.3 vs. 78.6). \ourmethod also consistently outperforms GroupViT, a method specialized for semantic segmentation.
The gains become more pronounced when we scale up to a larger model ViT-L/14, demonstrating good scalability of \ourmethod.

\begin{table}[!t]
% \vskip 0.2in
\caption{\textbf{Comparison with contrastive methods on vision-language tasks}: image captioning (CIDEr score) and VQA (accuracy). All methods are pretrained on YFCC15M with a ViT-B/16 visual encoder. SPARC uses the same extra synthetic text captions as \ourmethod. $\dagger$: our implementation.}
\label{tb:captioning_VQA}
\begin{center}
\vskip -0.1in
\resizebox{0.85\linewidth}{!}{
\begin{tabular}{lccc}
\toprule
 & Captioning & \multicolumn{2}{c}{VQA} \\
\cmidrule(lr){2-2}
\cmidrule(lr){3-4}
Method & COCO & GQA & VQAv2 \\
\midrule
CLIP~\citep{radford2021learning} & 108.3 & 44.6 & 55.2 \\
SPARC~\citep{SPARC}$\dagger$ & 109.7 & 45.4 & 56.2 \\
\ourmethod (ours) & \textbf{111.6} & \textbf{46.3} & \textbf{57.8} \\
\bottomrule
\end{tabular}
}
\end{center}
\vskip -0.2in
\end{table}

\begin{figure*}[!t]
\begin{center}
% \vskip -0.1in
\centerline{\includegraphics[width=0.83\linewidth]{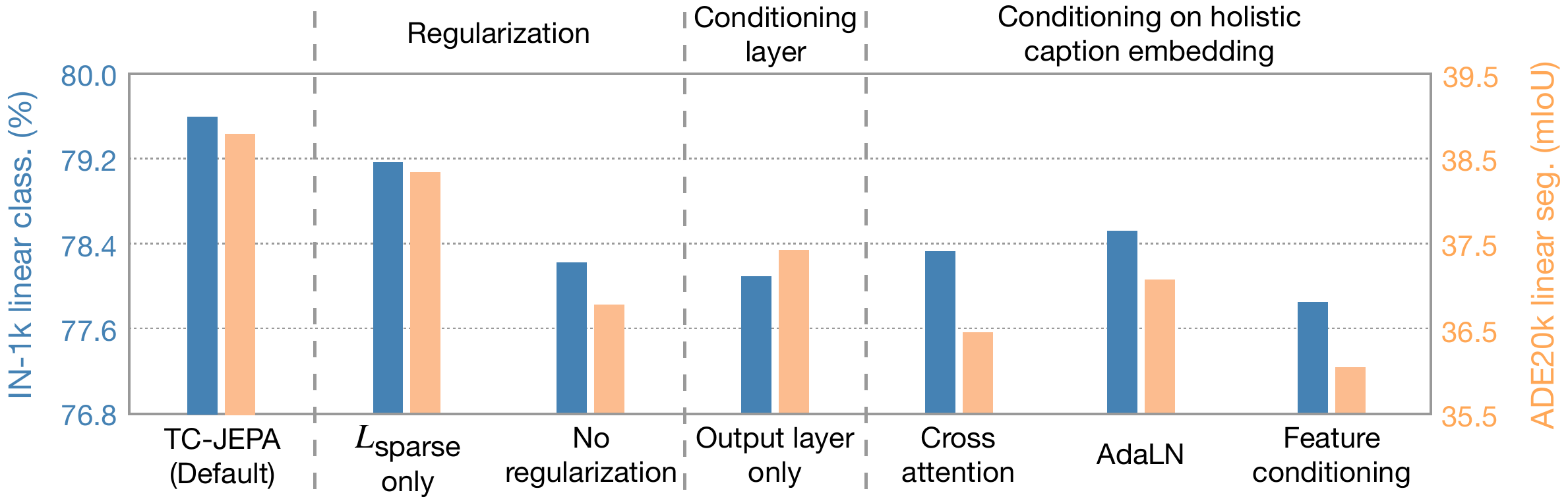}}
\end{center}
\vskip -0.25in
\caption{\textbf{Ablating the key components of our text conditioning method}. All baselines use the ViT-L/16 encoder pretrained on IN-1k (with the same synthetic text captions).} 
\label{fig:Ablations}
% \vskip -0.1in
\end{figure*}

\textbf{Vision-language tasks.} Recall the patch representations learned with \ourmethod are predictable as a function of text, encoding rich semantics in a predictive way. One hypothesis is that such predictive representations have strong multimodal capabilities, which could be better-suited than constrastively learned features for vision-language understanding and generation tasks. We empirically prove this hypothesis by evaluating the representations learned by \ourmethod and contrastive baselines on the Visual Question Answering (VQA) tasks on GQA~\citep{Hudson_2019_CVPR} and VQAv2~\citep{balanced_vqa_v2} datasets, as well as on the image captioning task on COCO dataset. Appendix~\ref{sec:eval_details} includes the evaluation details on these tasks that require fine-grained visual understanding and/or reasoning.

Table~\ref{tb:captioning_VQA} shows that \ourmethod outperforms the contrastive baselines CLIP and SPARC (with a fine-grained loss) on both captioning and VQA tasks. These results confirm the superior representation quality of our non-contrastive pretraining approach in the multimodal setting.

\subsection{Analysis}

\textbf{Ablations.} Fig.~\ref{fig:Ablations} ablates the key components of our text conditioning method, in order to disentangle their contributions from using the text supervision itself. We first observe a big performance drop when \ourmethod is trained without using the regularization terms $\mathcal{L}_{\text{sparse}}$ and $\mathcal{L}_{\text{consistency}}$. Adding $\mathcal{L}_{\text{sparse}}$ is particularly helpful because it helps to find useful patch-word similarities during text conditioning, leading to word-sensitive, semantic patch features that see a performance jump. $\mathcal{L}_{\text{consistency}}$ further improves by constraining the similarity consistency across different layers of the conditional feature predictor.
Second, it can be seen that conditioning a single layer of the predictor significantly underperforms multi-layer conditioning.

Finally, we compare our fine-grained conditioning method with popular conditioning methods that often use holistic text features (see method details in Appendix~\ref{sec:conditioning_ablation}). Fig.~\ref{fig:Ablations} shows the inferior performance of using cross attention and Adaptive Layer Normalization (AdaLN)~\citep{AdaLN,bar2025navigation}, both conditioned on the caption-level text embeddings to modulate predictor layers. Their performance gap with \ourmethod is especially apparent on the segmentation task. This is because conditioning at caption level, rather than word level, is unable to align patch features with fine-grained textual semantics, hence struggling with segmentation-like tasks that require local image understanding. Feature conditioning~\citep{Garrido2024LearningAL,zhou2025dinowm} the predictor inputs is further limited by single-layer conditioning, achieving the lowest task performance.

\begin{figure}[!t]
\begin{center}
% \vskip -0.1in
\centerline{\includegraphics[width=1.0\linewidth]{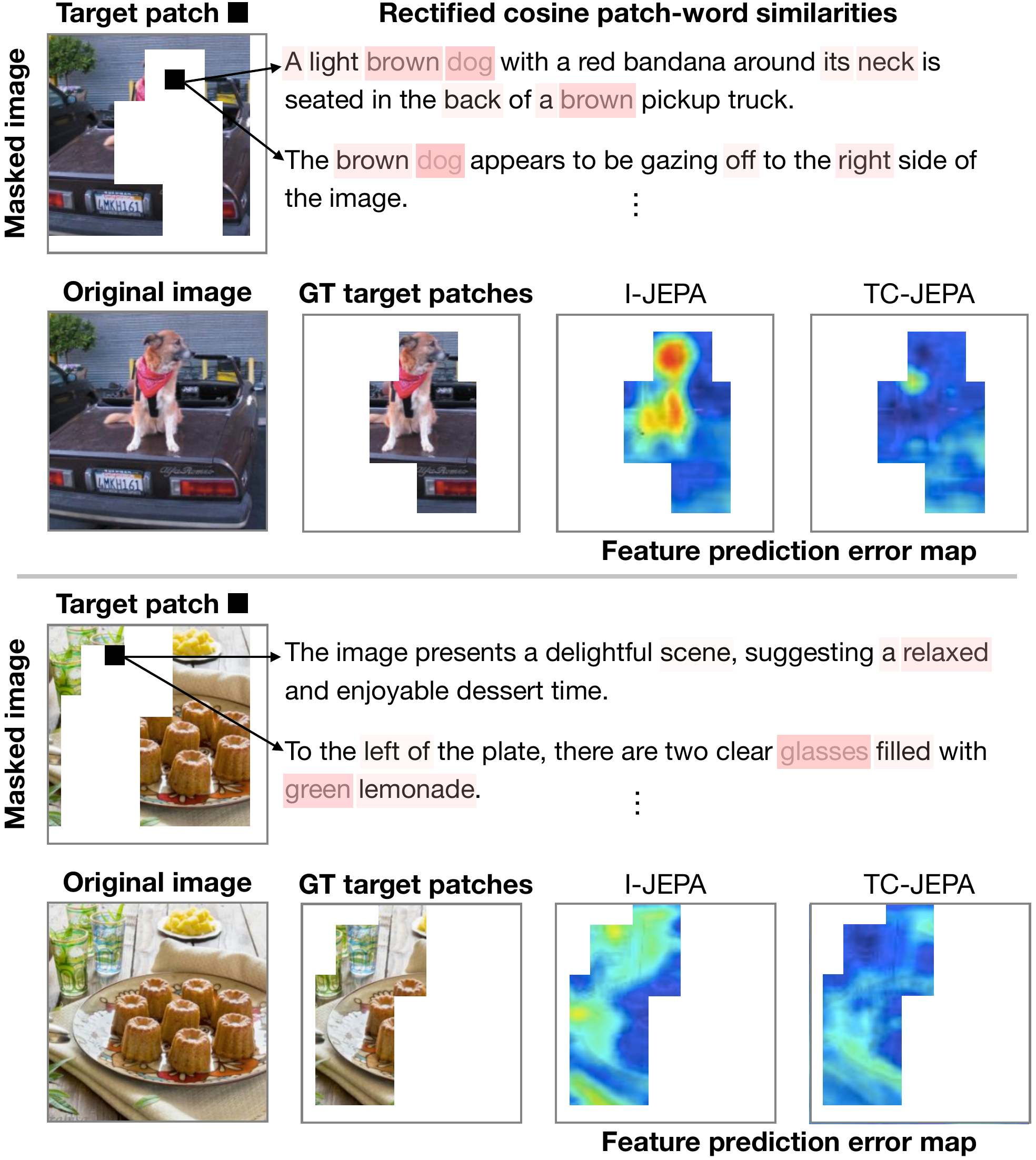}}
\end{center}
\vskip -0.1in
\caption{\textbf{Visualizing text-conditioned feature prediction}. We obtain sparse and semantic patch-word similarities (averaged across predictor layers) that are \textit{unsupervisedly} learned to aid target patch feature prediction.
This makes \ourmethod achieve lower feature prediction error than \ijepa, confirming that our text conditioner can indeed reduce prediction uncertainty.} 
\label{fig:visualization}
\vskip -0.2in
\end{figure}
% semantically meaningful/rich object-level, language-aligned, text-sensitive local patch features tokens-- high level semantic information-global image representations
% achieve strong alignment with LLMs/semantic alignment with language, obtained by pretraining with image-caption pairs. implicit image-text alignment
% language/fine-grained/multimodal understanding capabilities
% fine-grained spatial (patch) representations. capture spatial and semantic information

% mae/jepa retains fine-grained information about much of the local structure in the input image - reinforcing their role in capturing fine-grained localized image details

\textbf{Prediction visualization.} From Fig.~\ref{fig:visualization} we observe that 1) although our text-conditioned predictor identifies unsupervised patch-word correspondences, they are semantically meaningful. 2) Such fine-grained text conditioning helps reduce feature prediction uncertainty (\eg,~when predicting a dog sitting on a truck), leading to low prediction error, and more importantly, feature representations that are implicitly aligned with language.

\section{Conclusion}
In this paper, we introduce fine-grained text conditioning into the JEPA prediction task, which learns text-sensitive and semantically rich visual representations. We show our \ourmethod method can reduce feature prediction uncertainty and hence improve training stability. When evaluated on various downstream tasks, \ourmethod achieves strong performance with promising scaling properties. \ourmethod can be also viewed as a predictive vision-language pretraining approach, which compares favorably to contrastive ones especially on dense prediction and vision-language tasks.

\section*{Impact Statement}
This paper proposes a visual representation learning method based on text-conditional feature prediction within images. One potential societal impact is that, when the pretraining data of either the images or conditioning texts present (unintentional) biases, our method could inherit those biases in learned feature representations. As a result, one may observe unfair or discriminative outcomes that are sub-optimal for AI-assisted applications.

\bibliography{myref}
\bibliographystyle{icml2026}

%%%%%%%%%%%%%%%%%%%%%%%%%%%%%%%%%%%%%%%%%%%%%%%%%%%%%%%%%%%%%%%%%%%%%%%%%%%%%%%
%%%%%%%%%%%%%%%%%%%%%%%%%%%%%%%%%%%%%%%%%%%%%%%%%%%%%%%%%%%%%%%%%%%%%%%%%%%%%%%
% APPENDIX
%%%%%%%%%%%%%%%%%%%%%%%%%%%%%%%%%%%%%%%%%%%%%%%%%%%%%%%%%%%%%%%%%%%%%%%%%%%%%%%
%%%%%%%%%%%%%%%%%%%%%%%%%%%%%%%%%%%%%%%%%%%%%%%%%%%%%%%%%%%%%%%%%%%%%%%%%%%%%%%
\newpage
\appendix
\onecolumn
\section{Text Conditioning on Holistic Caption Embedding}
\label{sec:conditioning_ablation}

In the main paper, we introduce a text conditioning method based on the word embedding sequence $t=[t_1,\dots,t_S] \in \mathbb{R}^{d_t \times S}$ of a text caption sentence. In the literature, there are alternative conditioning methods that use the holistic caption embedding in various domains,~\eg,~\citep{lavoie2024modeling,Imagen}. The single vector representation of a caption is often obtained by average-pooling the word embeddings into $\bar t \in \mathbb{R}^{d_t}$. Here we list some strong conditioning baselines based on $\bar t$. Note compared to $\bar t$-based conditioning methods, conditioning on word sequence $t$ has one key advantage: one can model the fine-grained correspondences between image patches and word tokens, akin to visual grounding. Such text-conditioned patch representations can thus play a better role in capturing fine-grained and semantic image details.

\textbf{Cross attention baseline.} As a direct comparison, we implement cross attention over $\bar t \in \mathbb{R}^{d_t}$ at each layer. Following Eq.~(\ref{eq2}) and assuming one caption per image, we now have $\text{Attention}(q^{(l)},K^{(l)},V^{(l)})$ with:
\begin{equation}
q^{(l)} =W^{(l)}_Q \cdot q, \;\; K^{(l)}=W^{(l)}_K \cdot \bar t, \;\; V^{(l)}=W_V^{(l)} \cdot \bar t,\\
\label{eq5}
\end{equation}
where $W^{(l)}_Q \in \mathbb{R}^{d \times d}$, $W^{(l)}_K \in \mathbb{R}^{d \times d_t}$ and $W^{(l)}_V \in \mathbb{R}^{d \times d_t}$ are the learnable matrices at the $l^{th}$ layer. Note cross-attention is now computed between the image patch and entire caption sentence (not words). Therefore, this baseline can no longer capture the fine-grained patch-word correspondences, and we discard the sparsity $\mathcal{L}_{\text{sparse}}$ and consistency $\mathcal{L}_{\text{consistency}}$ constraints that are designed for regularizing patch-word similarities. In case of multiple captions with their caption embeddings $\{\bar t^n\}_{n \in [1,N]}$, we similarly max-pool the independently conditioned patch features at each layer.

\textbf{Adaptive Layer Normalization (AdaLN).} AdaLN~\citep{AdaLN,bar2025navigation} provides an efficient way to text condition the predictor $g_{\phi}$ on aggregated $\bar t$. Specifically, we feed $\bar t$ to an AdaLN block to generate scale and shift coefficients that modulate the LayerNorm outputs of each predictor layer. $\{\bar t^n\}$ of different captions produce different modulation outputs at each layer, which is max-pooled again. The parameters of the AdaLN block are jointly learned with that of the predictor.

\textbf{Feature conditioning.} Feature conditioning is frequently used as a simple conditioning method in the literature~\citep{Garrido2024LearningAL,zhou2025dinowm,baldassarre2025dinoworld}. The idea is to add $\bar t$ as extra dimensions to the predictor input,~\ie,~$z_{x_i}$ for $i\in B_x$ and mask token $m_j$ for $j\in B_y$. Formally, the input features are updated as:
\begin{eqnarray}
&&z_{x_i} \leftarrow z_{x_i}+\text{MLP}\left(\text{LayerNorm} \left ([z_{x_i},\bar t\,] \right) \right), \notag\\
&&m_j \leftarrow m_j+\text{MLP}\left(\text{LayerNorm} \left ([m_j,\bar t\,] \right) \right),
\label{eq6}
\end{eqnarray}
where the concatenated features (after LayerNorm) are fed into an MLP and then added back to the mask token (residual connection). To handle multiple text feature vectors $\{\bar t^n\}$, we first perform feature conditioning with each $\bar t^n$ using Eq.~(\ref{eq6}), and then max pool the multiple conditioned input features.

Despite the simplicity, feature conditioning suffers from several drawbacks: 1) The increased feature dimensions require a larger model capacity for the predictor, which incurs non-negligible overhead in memory and compute. 2) Directly mixing features of different types (textual $\bar t$ vs. image features $z_{x_i}$ vs. positional mask token $m_j$) complicates training. 3) The conditioning only happens at the predictor's input level, and its influence can rapidly diminish as depth increases.

\section{Synthetic Image Captions}
\label{sec:gen_cap}

We follow~\citep{DreamLIP} and query ShareGPT4V~\citep{chen2024sharegpt4v} to generate image captions in a scalable way. Specifically, ShareGPT4V is queried with two prompts: 1) ``Describe the image in short'' that often generates succinct text descriptions in 1 to 2 captions (sentences) for a given image. 2) ``Describe the image in detail'' that generates a long list of detailed captions, each one often focusing on a different visual aspect.
Note the text captions generated from the two prompts are all truncated or padded to the same length. Then we simply combine all these generated captions for each image, and treat them as synthetic image captions altogether. Fig.~\ref{fig:statistics_cap} shows the number of synthetic caption sentences per image for the four considered datasets in this paper.

\begin{figure*}[!t]
\begin{center}
% \vskip -0.07in
\centerline{\includegraphics[width=1.0\linewidth]{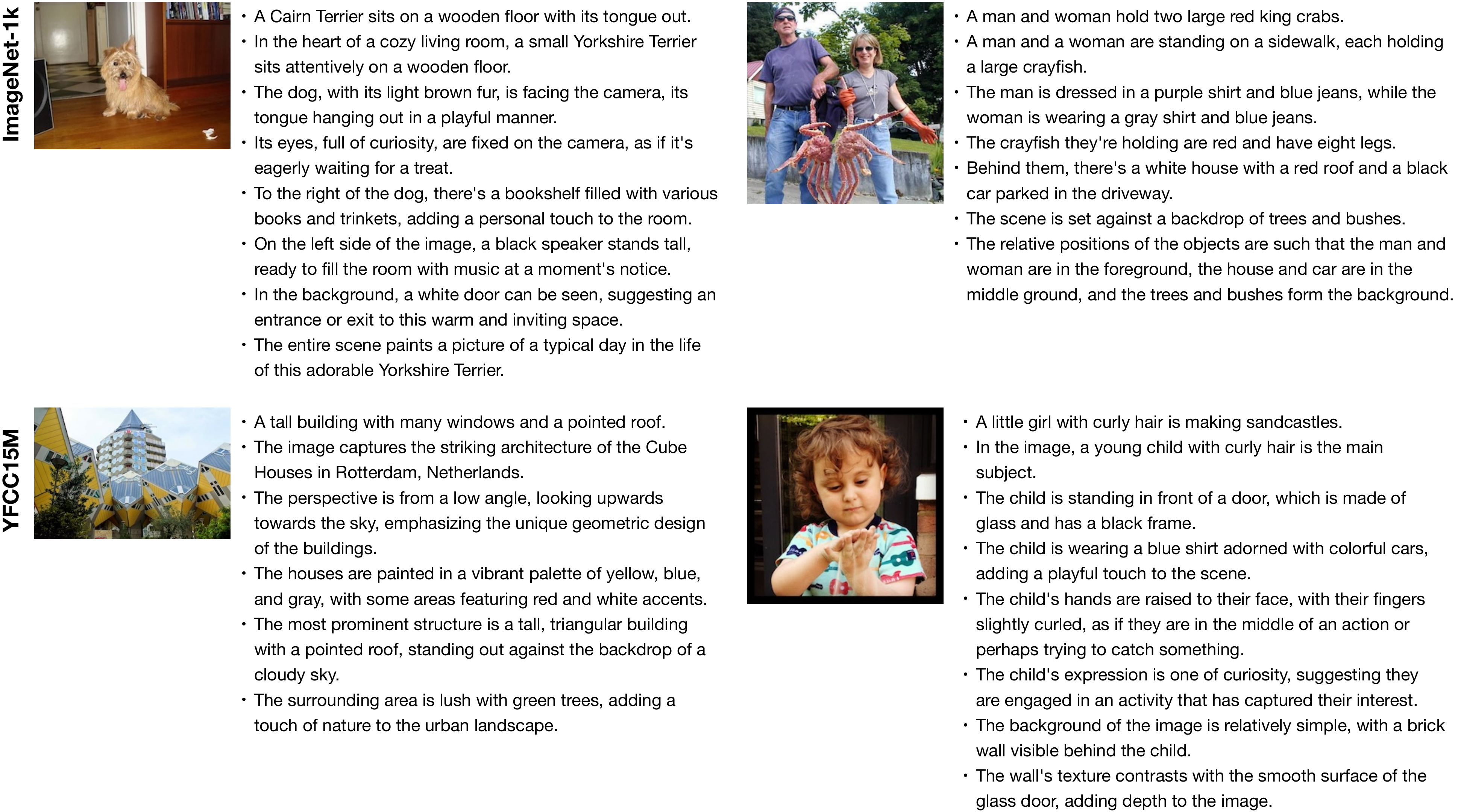}}
\end{center}
\vskip -0.1in
\caption{Example synthetic image captions for IN-1k and YFCC15M datasets.} 
\label{fig:example_cap}
\vskip 0.1in
\end{figure*}

\begin{figure*}[!t]
\begin{center}
% \vskip -0.07in
\centerline{\includegraphics[width=1.0\linewidth]{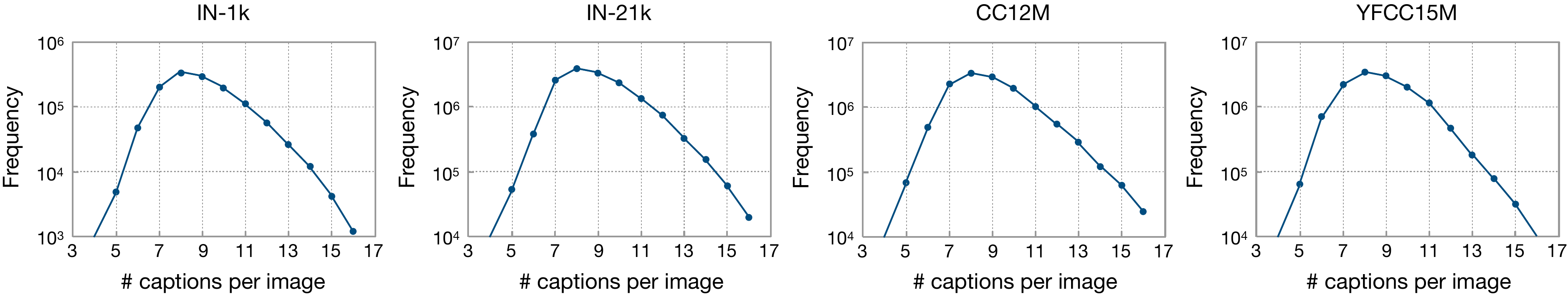}}
\end{center}
\vskip -0.1in
\caption{Statistics of image captions (sentences) synthesized on different datasets.} 
\label{fig:statistics_cap}
% \vskip 0.1in
\end{figure*}

Fig.~\ref{fig:example_cap} exemplifies the synthetic image captions on IN-1k and YFCC15M datasets. We can see that different captions often capture different visual properties of the image, including but not limited to: 1) scene composition, 2) visual attributes of both foreground and background objects, and 3) spatial relationships or interactions between objects. These diverse text captions can model the richness of visual input, hence are suitable to guide the difficult JEPA prediction task at different image locations.
Note there may be hallucinations in the generated captions. We rely on the attention mechanism in our text conditioning method to filter out noisy and image-irrelevant information in the generated captions.

\section{Training Details and Evaluations}
\label{sec:train_eval_details}
\subsection{Pretraining Details on ImageNet}
\label{sec:IN_train_details}

\textbf{Architectures.} For the image encoder $f_{\theta}$, we use standard Vision Transformer (ViT) architectures of varying capacities: ViT-B/16, ViT-L/16 and ViT-H/14. The predictor $g_{\phi}$ is a narrow ViT, with fixed embedding dimension 384 and the number of self-attention heads equal to that of $f_{\theta}$. The predictor depth is 6 for smaller encoder (ViT-B/16), or 12 for larger encoders ViT-L/16 and ViT-H/14. In either case, every predictor layer is text-conditioned by a light residual cross-attention layer, which is followed by a two-layer MLP network and LayerNorm.

\textbf{Masking strategy} plays a key role for JEPA methods to learn semantic representations. We use the multi-block masking strategy in \ijepa~\citep{assran2023self} to sample 1 context and 4 target blocks within each image using the same masking hyper-parameters.

\textbf{Optimization.} For pretraining on IN-1k, we use the AdamW optimizer~\citep{loshchilov2018decoupled} to jointly optimize the encoder, predictor and text conditioner. We generally follow the recipe of~\citep{assran2023self} including the fixed batch size 2048, max learning rate $10^{-3}$ with a warmup and then cosine decay schedule, and weight decay linearly increased from 0.04 to 0.4. Tuning the learning rate and weight-decay schedules does not bring much benefit in our experiments. Instead, we found our properly regularized \ourmethod objective makes JEPA learning robust when conditioned by text.
We similarly train for 600 epochs for ViT-B/16 and ViT-L/16 encoders, and 300 epochs for ViT-H/14 encoder. All models are trained at resolution $224\times 224$ pixels.

For pretraining on IN-21k, we follow similar configs as mentioned above, except that we train the ViT-L/16 encoder for the equivalent of 1200 IN-1k epochs, and the ViT-H/14 encoder for the equivalent of 900 IN-1k epochs.

\subsection{Pretraining Details on Image-Text Datasets}

\textbf{Optimization.} We further scale up the pretraining dataset and use the image-text dataset YFCC15M and the combined mixture of CC12M+YFCC15M. We follow similar configs for ImageNet pretraining as detailed in Section~\ref{sec:IN_train_details}, including the hyperparameters of batch size, learning rate and weight-decay. All models are trained at resolution $224\times 224$ pixels. One modification we made is on the training epochs to accommodate the increased dataset size. Specifically, on YFCC15M or the combined CC12M+YFCC15M, we train the ViT-B/16 and ViT-L/14 image encoders for the same 50 epochs. This is close to the training schedules adopted for many compared vision-language models to enable fair comparisons.
We adjust our warmup epochs to 10 accordingly.

\subsection{Evaluations}
\label{sec:eval_details}

\textbf{Linear classification}. We use the exact linear probing recipes in~\citep{assran2023self} for evaluations on each of the classification datasets: IN-1k, CIFAR100, Places205 and iNat18. Concretely, we adopt the evaluation protocol from VISSL~\citep{goyal2021vissl} for linear probing with a frozen backbone. When evaluating methods like iBOT, DINO and MAE, their \texttt{[cls]} token representations are used for evaluation. While for methods of \ijepa, StoP and our \ourmethod, they are pretrained without a \texttt{[cls]} token. We use the target encoder of these methods, and utilize the average-pooled patch representation for linear evaluation.

\textbf{Object detection.} We follow the experimental protocol of~\citep{zhou2021ibot} to evaluate the detection performance on COCO dataset with different ViT backbones: The ViT feature maps are scaled to 4 different sizes to be used in FPN (Feature Pyramid Network). Then Mask R-CNN~\citep{maskrcnn} is fine-tuned with 1$\times$ schedule for 12 epochs, using the same fine-tuning hyperparameters.

\textbf{Semantic segmentation.} We consider the two setups in~\citep{zhou2021ibot,bao2022beit} for evaluation on ADE20k and Pascal VOC datasets. First is the \textit{linear evaluation} protocol for an explicit measurement of patch representation quality. We train a linear classifier on top of the patch features of a frozen backbone to predict class logits. AdamW optimizer is used for hyperparameter sweep over the learning rate and weight decay. Second, we \textit{fine-tune} the ViT model end-to-end using an UperNet~\citep{xiao2018unified} segmentation head. We fine-tune on 512$\times$512 resolution for 160k iterations using the suggested recipe (batchsize, learning rate and stochastic depth,~\etc). No multi-scale training and testing is used.

\textbf{Vision-language tasks.} We evaluate the multimodal understanding capabilities of learned representations on image captioning and VQA tasks. For this, we freeze the visual encoder and train a LiT-Decoder~\citep{LiTDecoder} on top in a multi-task setup. Concretely, a single 12-layer autoregressive decoder is trained on the frozen encoder, which learns a multi-task model for captioning and VQA. We carefully follow the official implementation and similarly use a unified image preprocessing across tasks. To ease multi-task training on different task data (COCO, GQA and VQAv2), we also use the task mixing strategies and task prompts for decoder conditioning. The same training hyperparameters are used, including the learning-rate, weight-decay, epochs, label smoothing and dropout.

For comparing with different representation learning methods, we simply replace the respective visual encoder under the LiT-Decoder framework. At inference time, we use greedy decoding for the VQA tasks and beam search with 4 beams for captioning. The VQA tasks are evaluated in terms of accuracy, with only exact string matches counted as correct. We evaluate captioning performance in CIDEr score~\citep{VedantamZP15}.

\section{Training Efficiency}
\label{sec:train_efficiency}

The compute cost of \ijepa baseline is dominated by the forward pass of image encoders $f_{\theta}$ and $f_{\bar\theta}$ to produce the context and target patch representations. Feature prediction and loss computation is only based on a lightweight predictor $g_{\phi}$. Note our \ourmethod text conditions the predictor $g_{\phi}$ rather than bulky encoders. Hence the computational overhead introduced during pretraining is marginal. Also, text conditioning does not impact the inference stage, since only the encoder $f_{\theta}$ is used at test time while the predictor and text-conditioner are both discarded.

\begin{figure*}[!t]
\begin{center}
% \vskip -0.07in
\centerline{\includegraphics[width=1.0\linewidth]{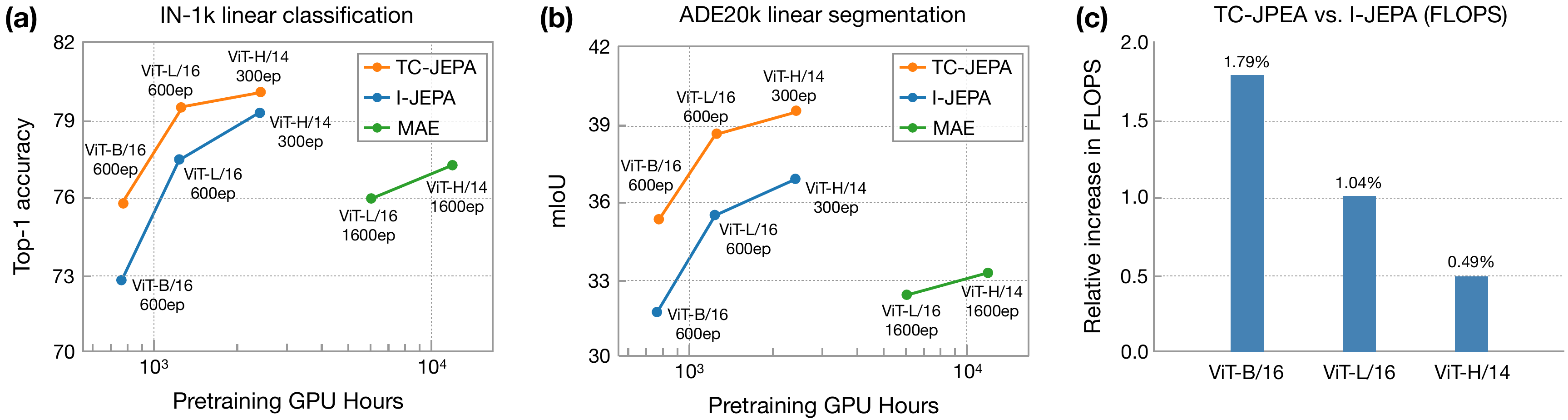}}
\end{center}
\vskip -0.2in
\caption{\textbf{Model efficiency analysis}. \textbf{(a-b)} Downstream performance vs. pretraining GPU hours on IN-1k. \textbf{(c)} Relative increase in FLOPS when comparing our \ourmethod to \ijepa baseline with increased encoder size.} 
\label{fig:compute_cost}
% \vskip -0.15in
\end{figure*}

Fig.~\ref{fig:compute_cost} highlights the training efficiency of \ourmethod using the example of IN-1k pretraining, when $N=8$ text captions are used for each image. As can be seen from (a-b), \ourmethod’s training time is only slightly higher than \ijepa for the same model size and training epochs, while having notable improvements on linear classification and segmentation performance. Interestingly, a large \ourmethod model (ViT-L/16) outperforms the huge \ijepa model (ViT-H/14) while requiring a significantly lower training time. Note the increased training time with text conditioned $g_{\phi}$ will become negligible as we scale up the size of encoder $f_{\theta}$. The same conclusion may be also made with respect to the FLOPS --- see (c) for the relative increase in FLOPS when comparing \ourmethod to \ijepa baseline with increased encoder size. 
Finally in (a-b), when compared to the pixel reconstruction method MAE, \ourmethod converges in roughly 5$\times$ fewer epochs, achieving significant compute savings and performance gains at the same time.

\begin{table}[!t]
\vskip 0.1in
\caption{\textbf{Multi-caption fusion strategy}: MaxPool (default) vs. alternatives when pretraining the ViT-L/16 encoder on IN-1k.}
\label{tb:multi_caption_fusion}
\begin{center}
\vskip -0.1in
\resizebox{0.55\linewidth}{!}{
\begin{tabular}{lccc}
\toprule
Fusion strategy & AvgPool & AttentionPool & MaxPool\\
\midrule
IN-1k linear classification & 79.3 & 79.5 & \textbf{79.6}\\
ADE20k linear segmentation & 38.4 & \textbf{39.0} & 38.8\\
\bottomrule
\end{tabular}
}
\end{center}
% \vskip -0.2in
\end{table}

\section{Additional Ablations}
\label{sec:additional_ablation}

\textbf{Multi-caption fusion strategy}. Table~\ref{tb:multi_caption_fusion} compares different strategies to fuse the patch features conditioned by multiple text captions. We see that MaxPool consistently outperforms AvgPool since the former can better select the most useful text information for conditioning purposes. The AttentionPool strategy is on par with or slightly better than MaxPool, but is more costly with the extra parameters learned for attention pooling at every predictor layer. Hence MaxPool is chosen as our default fusion strategy due to its good performance-cost trade-off.

\begin{figure*}[!t]
\begin{center}
% \vskip -0.07in
\centerline{\includegraphics[width=1.0\linewidth]{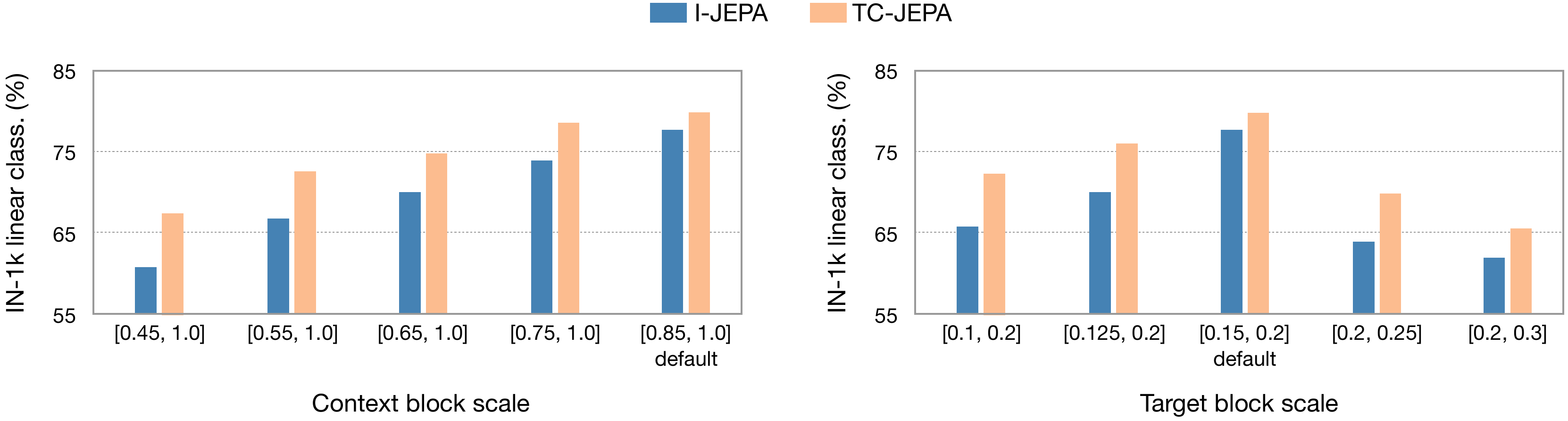}}
\end{center}
\vskip -0.2in
\caption{\textbf{Training stability across different ranges of the context and target block scale} used for multi-block masking during pretraining. Ablations are performed by pretraining the ViT-L/16 encoder on IN-1k.} 
\label{fig:Masking_hyperparas}
% \vskip -0.15in
\end{figure*}

\textbf{Training stability analysis}. We conduct this analysis via ablating the key hyper-parameters of the multi-block masking strategy used for pretraining --- the context and target block scale. As mentioned in~\citep{assran2023self}, the sampled context/target blocks for feature prediction could affect the semantic level of representations learned by the JEPA-style methods.
Fig.~\ref{fig:Masking_hyperparas} shows the ablation results across varying ranges of the context/target block scale. We use the IN-1k linear probing performance as a measure of representation quality. It can be seen that our \ourmethod is much less sensitive to a wide range of block sizes than \ijepa. This suggests that text conditioning makes JEPA methods more robust to learn semantic representations.

\begin{figure*}[!t]
\begin{center}
\vskip 0.2in
\centerline{\includegraphics[width=1.0\linewidth]{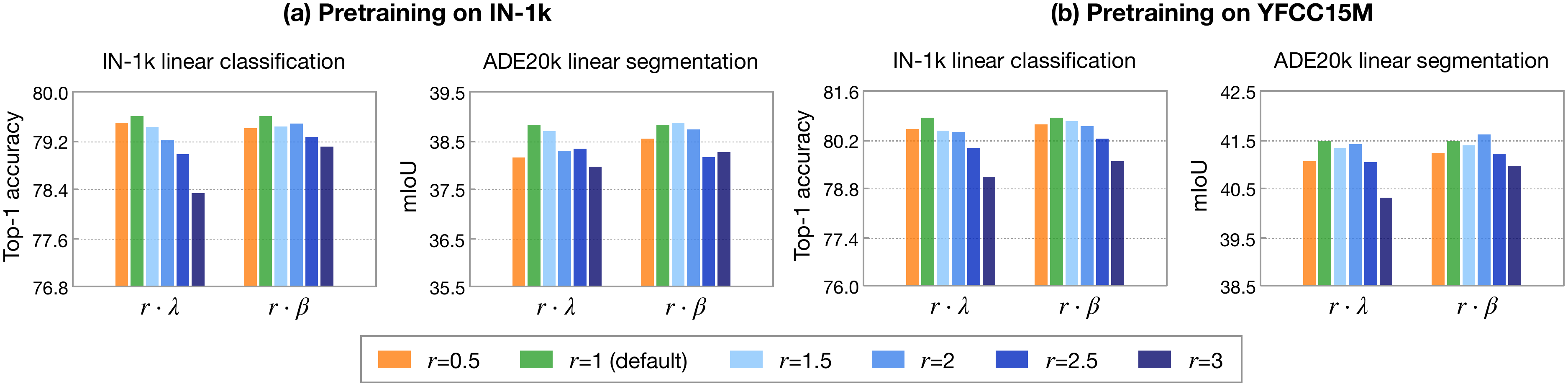}}
\end{center}
\vskip -0.15in
\caption{\textbf{Sensitivity analysis of the loss coefficients $\lambda$ and $\beta$ in Eq.~(\ref{eq4})}. We sweep over the loss coefficients by applying a varying multiplier $r$ to them.
Then we perform sensitivity analysis for pretraining the ViT-L/16 encoder on \textbf{(a)} IN-1k as well as \textbf{(b)} the image-text dataset YFCC15M (evaluated on two tasks). Stable convergence is observed when $\lambda$ and $\beta$ are scaled by $r\in[0.5,2.5]$, with the exact values picked for $\lambda$ and $\beta$ having limited influence on the final performance.} 
\label{fig:loss_hyperparas}
\vskip 0.1in
\end{figure*}

\begin{figure*}[!t]
\begin{center}
\vskip 0.15in
\centerline{\includegraphics[width=1.0\linewidth]{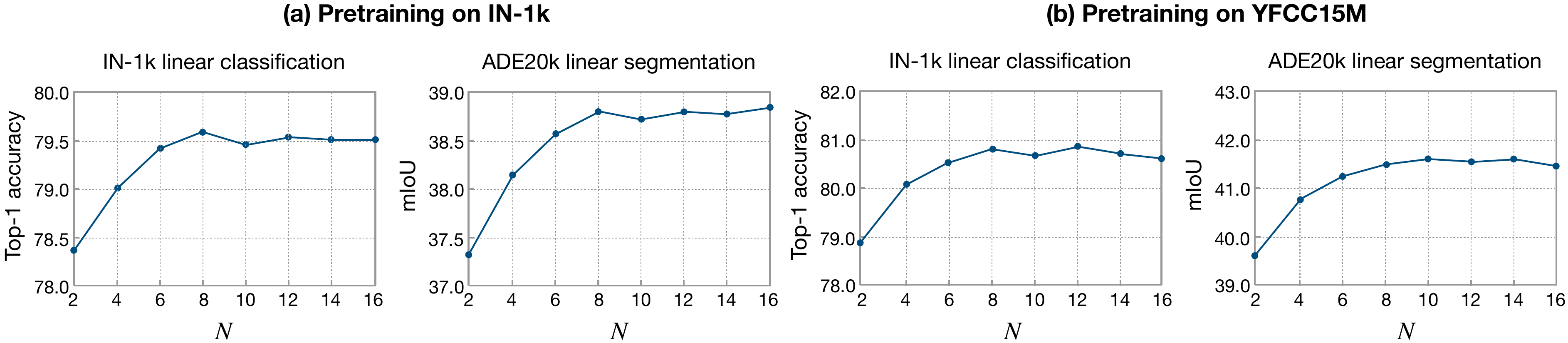}}
\end{center}
\vskip -0.2in
\caption{\textbf{Sensitivity analysis of $N$, the number of randomly sampled text captions for \ourmethod training}. We perform the sensitivity analysis when pretraining the ViT-L/16 encoder on \textbf{(a)} IN-1k as well as \textbf{(b)} the image-text dataset YFCC15M (evaluated on two tasks). The default $N$ is set to 8, after which performance saturates.} 
\label{fig:N_hyperpara}
% \vskip -0.15in
\end{figure*}

\textbf{Sensitivity analysis of hyperparameters}. Fig.~\ref{fig:loss_hyperparas} shows the ablation results on the loss coefficients $\lambda$ and $\beta$, when pretraining the ViT-L/16 encoder on either IN-1k or the image-text dataset YFCC15M. Performance is found robust to a wide range of loss coefficients.

Fig.~\ref{fig:N_hyperpara} shows the ablation results on the number of text captions $N$ used for text conditioning during \ourmethod training (on IN-1k or YFCC15M dataset using the ViT-L/16 encoder). We observe that performance saturates near $N=8$, which is set as our default value for a good trade-off between performance and compute cost.

\begin{figure*}[!t]
\begin{center}
% \vskip 0.15in
\centerline{\includegraphics[width=0.6\linewidth]{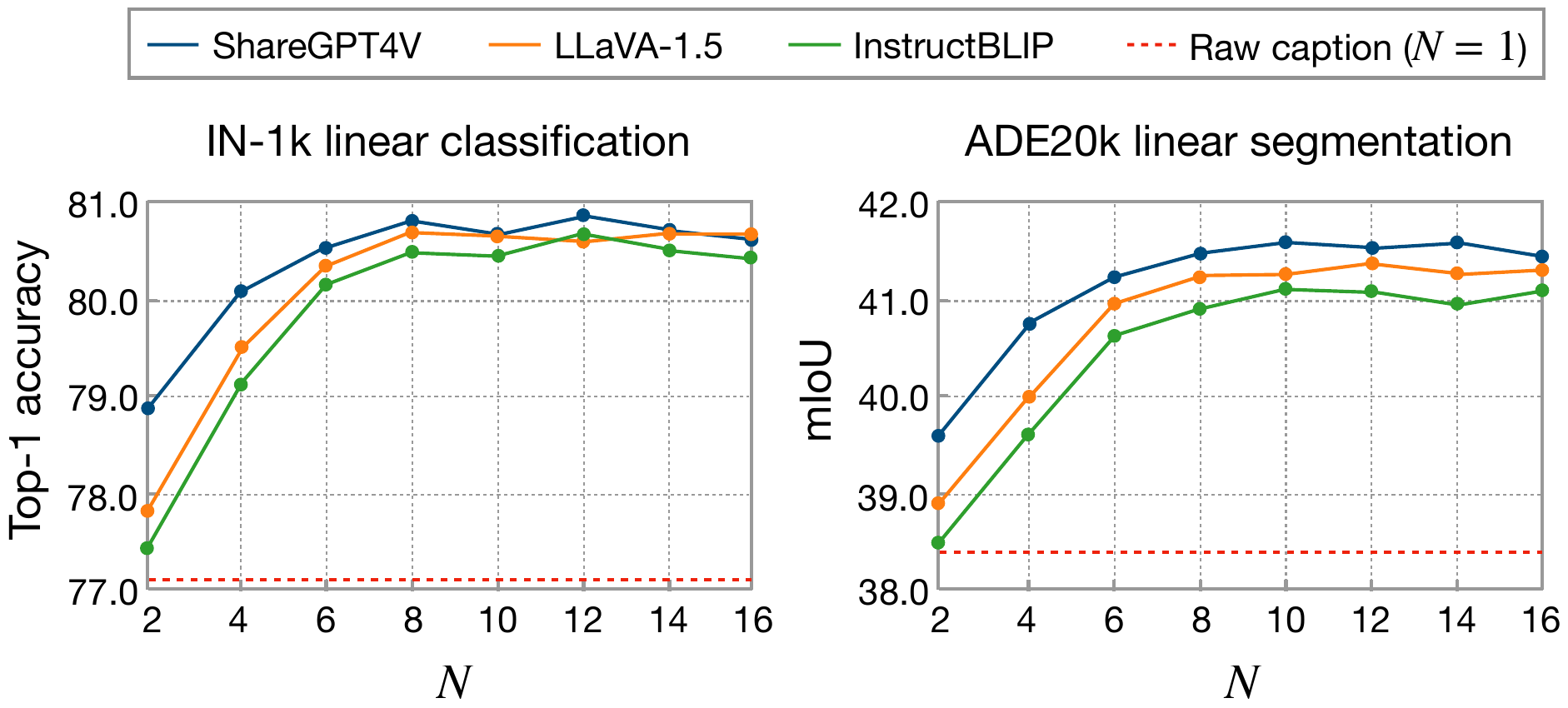}}
\end{center}
\vskip -0.2in
\caption{\textbf{Robustness to synthetic caption quality}. We train the ViT-L/16 encoder on the YFCC15M dataset, which is enriched with synthetic captions of varying quality generated by different models (ShareGPT4V, LLaVA-1.5 and InstructBLIP). We compare downstream performance as a function of synthetic caption quantity $N$. We also compare with the baseline that uses only the $N=1$ raw caption from YFCC15M for \ourmethod training.} 
\label{fig:Caption_model_analysis}
% \vskip -0.15in
\end{figure*}

\textbf{Robustness to synthetic caption quality}. In the main paper, we use ShareGPT4V to synthesize image captions for \ourmethod training. Following~\citep{DreamLIP}, we also experiment with using two other models to synthesize captions, based on LLaVA-1.5~\citep{liu2023improvedllava} and InstructBLIP~\citep{dai2023instructblip}. These models generate captions of different quality and styles, and their outputs are usually shorter or less descriptive than those from the default ShareGPT4V model. Fig.~\ref{fig:Caption_model_analysis} compares their text conditioning effects when pretraining ViT-L/16 encoder on the enriched YFCC15M dataset with varying $N$, the number of randomly sampled synthetic captions.

We observe that ``weaker'' captioning models lag far behind ShareGPT4V with small $N$ (2 or 4), but significantly narrow the gap when $N$ reaches 8 (default value) or larger. Note large $N$ means there are more diverse captions to potentially cover different visual aspects that we can cross-attend to during text conditioning. In other words, our observations suggest that caption quantity/diversity matters most; and \ourmethod is reasonably robust to the choice of captioning models and their captioning quality/style, as long as they can generate a sufficient number of captions with large diversity. When there are more than enough captions (e.g., $N>8$) likely with noticeable noise or hallucinations, our text conditioner is able to filter out the noisy and irrelevant information via attention mechanism.

Furthermore, Fig.~\ref{fig:Caption_model_analysis} compares with the baseline that uses the raw, human-annotated caption from YFCC15M ($N=1$). Results confirm the benefits of using diverse synthetic captions over short human annotations for text conditioning purposes.

%%%%%%%%%%%%%%%%%%%%%%%%%%%%%%%%%%%%%%%%%%%%%%%%%%%%%%%%%%%%%%%%%%%%%%%%%%%%%%%
%%%%%%%%%%%%%%%%%%%%%%%%%%%%%%%%%%%%%%%%%%%%%%%%%%%%%%%%%%%%%%%%%%%%%%%%%%%%%%%

\end{document}

%% file: math_commands.tex
\usepackage{amsmath,amsfonts,bm}